\documentclass[lettersize,journal]{IEEEtran}
\usepackage{amssymb,amsmath,amsfonts}
\usepackage{algorithmic}
\usepackage{algorithm}
\usepackage{array}
\usepackage{booktabs}
\usepackage{colortbl}
\usepackage[hidelinks]{hyperref}
\usepackage{graphicx}
\usepackage{makecell}
\usepackage[mathlines, switch]{lineno}
\usepackage{multirow}
\usepackage[sort&compress, numbers]{natbib}
\usepackage{pifont}
\usepackage{stfloats}
\usepackage[caption=false,font=footnotesize]{subfig}
\usepackage{textcomp}
\usepackage{url}
\usepackage{verbatim}
\usepackage[dvipsnames]{xcolor}
\usepackage{subcaption}
\usepackage{tabularx}
\usepackage{placeins}

\hypersetup{
  colorlinks = true,
  urlcolor = CadetBlue,
  linkcolor = Cerulean,
  citecolor = Maroon
}

\definecolor{whitesmoke}{rgb}{0.96, 0.96, 0.96}

\begin{document}

\title{A Diffusion-Refined Planner with Reinforcement Learning Priors for Space-Constrained Parking}

\author{
Mingyang~Jiang, Yueyuan~Li, Jiaru~Zhang, Songan~Zhang, and~Ming~Yang%
\thanks{This work was supported in part by the National Natural Science Foundation of China under Grant 62573289, Grant 52402504, and Grant U22A20100. \textit{(Corresponding author: Ming Yang; e-mail: MingYANG@sjtu.edu.cn.)}}%
\thanks{Mingyang Jiang, Yueyuan Li, and Ming Yang are with the School of Automation and Intelligent Sensing, Shanghai Jiao Tong University, Shanghai 200240, China.}%
\thanks{Jiaru Zhang is with the School of Computer Science, Shanghai Jiao Tong University, Shanghai 200240, China.}%
\thanks{Songan Zhang is with the Global Institute of Future Technology (GIFT), Shanghai Jiao Tong University, Shanghai 200240, China.}%
}

\maketitle

\begin{abstract}
Automated parking is an important function of intelligent vehicles and smart transportation systems, particularly in dense urban environments where parking spaces are limited and low-speed maneuvers must be executed safely and precisely. 
In space-constrained parking scenarios, high-precision maneuvers are required to achieve a high success rate in planning, yet existing approaches still face difficulties in capturing the precise and complex action distributions required in such settings. 
In this paper, we propose \textbf{DRIP}, a \textbf{D}iffusion-\textbf{R}ef\textbf{I}ned \textbf{P}lanner for automated parking, which leverages reinforcement learning (RL) prior action distributions to guide trajectory generation. Specifically, an RL-pretrained policy provides an informative coarse prior over feasible parking actions, and a diffusion-based denoising process progressively refines this prior into a more precise action distribution. We further introduce a prior-aligned training procedure that aligns the denoising trajectory with the RL prior, improving the consistency between training and inference and enabling effective refinement under tight spatial constraints. 
We evaluate DRIP across parking scenarios with varying degrees of spatial constraints. Experimental results show that DRIP improves planning success rates by 10.4\% over strong baselines while producing higher-quality trajectories, demonstrating its effectiveness for space-constrained automated parking planning.

\end{abstract}


\section{Introduction}

The growth of vehicle ownership in urban areas has intensified parking demand, while the supply of parking spaces remains constrained by limited urban land and infrastructure capacity~\cite{ma2024urbanparkingreview}. 
As urban parking has become increasingly integrated with transportation planning, the imbalance between parking supply and demand is widely recognized as a persistent issue affecting urban mobility, land use, and traffic operation~\cite{mingardo2015urban}. 
Under these supply-demand and land-use pressures, improving the utilization of existing parking facilities has become increasingly important, which in turn raises the need to handle parking layouts and maneuvering conditions with tighter spatial margins. 
The available clearance for local parking maneuvers is further reduced by the increasing size of passenger vehicles. 
Vehicle-market statistics show that the average car width in the EU increased from 170.5~cm in 2001 to 180.2~cm in 2020, and 52\% of the top 100 new models in 2023 exceeded the 180~cm minimum width commonly specified for on-street parking in major cities~\cite{transportenvironment2024wider}. 
Consistent with this trend, an industry analysis based on UK parking-bay dimensions reported that an average top-selling car occupied 56.2\% of a standard parking bay in 1976, increasing to 70.6\% in 2024~\cite{motorpoint2025parking}. 
Together, the pressure to use parking space more efficiently and the reduced maneuvering clearance make reliable automated parking planning increasingly important. 
Such planning can reduce low-speed collision risks, decrease unnecessary maneuvers, and improve parking efficiency~\cite{khalid2021smart}.

\begin{figure}[t]
  \centering
  \subfloat[]{\includegraphics[width=0.16\textwidth]{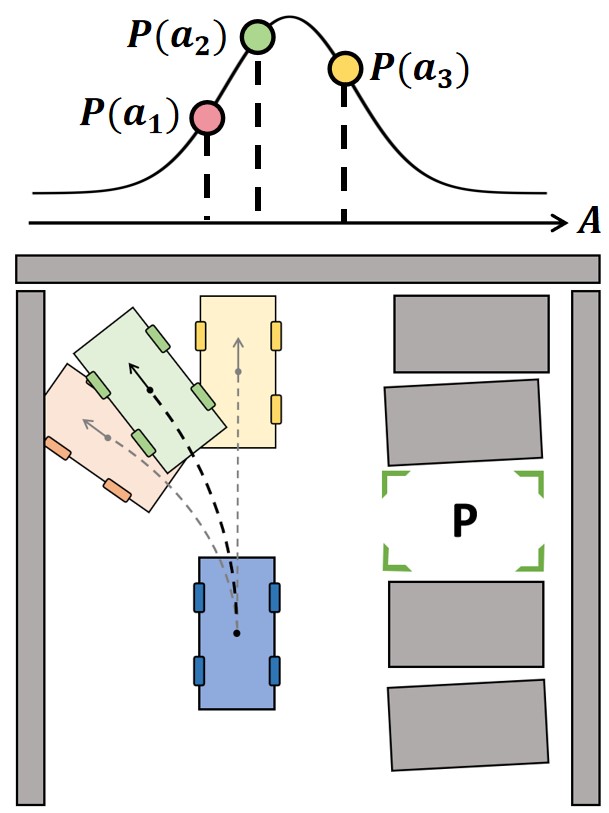}}\hfill
  \subfloat[]{\includegraphics[width=0.16\textwidth]{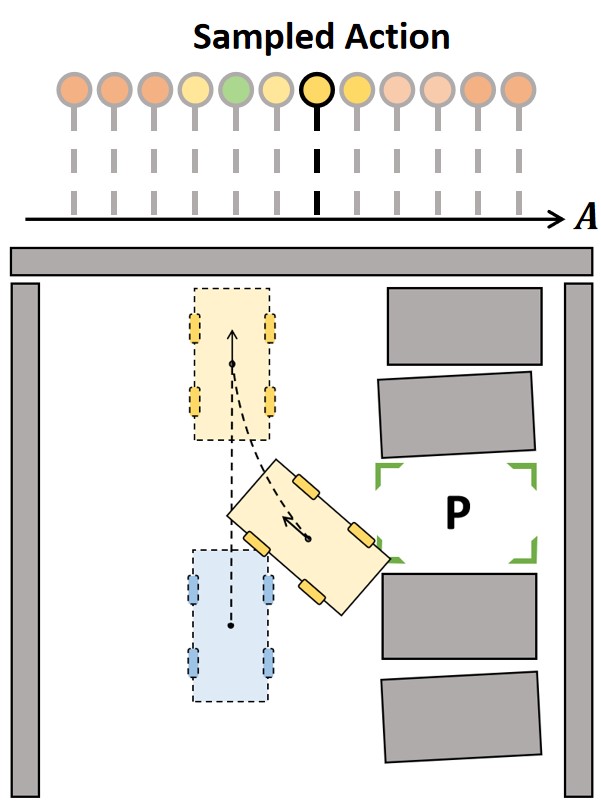}}\hfill
  \subfloat[]{\includegraphics[width=0.16\textwidth]{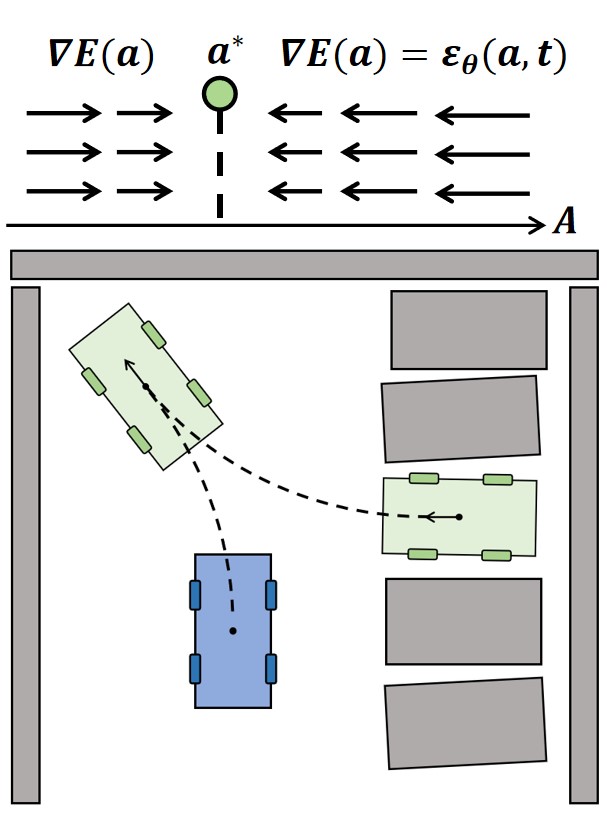}}\hfill
\caption{Comparison between action distribution prediction and diffusion-based action generation for parking.
(a) A learning-based planner parameterizes a conditional action distribution (or probabilistic policy) and selects actions by sampling from the distribution or using a deterministic action with the highest probability.
This leaves non-negligible probability mass on suboptimal maneuvers (b), which, in constrained spaces, may lead to collisions or additional corrective steps.
(c) A diffusion policy generates actions implicitly via learning a gradient field and iteratively refines noisy proposals toward feasible, high-precision trajectories.}

  \label{fig: intro}
  \vspace{5pt}
\end{figure}

However, planning in space-constrained parking spaces remains challenging, as the planner must generate high-precision maneuvers under non-holonomic constraints.
Traditional rule-based or optimization-based methods have been widely studied and deployed in relatively simple scenarios \cite{kim2014auto,sedighi2019guided}. 
Their performance can degrade in complex environments where the planner must reason over narrow feasible regions and diverse obstacle configurations \cite{likmeta2020combining}. 
This has motivated learning-based approaches that formulate parking path planning as a sequential action or trajectory prediction problem, learning data-driven mappings from environmental observations to driving actions \cite{teng2023motion}. 
Despite their flexibility, many learning-based planners predict actions directly, either through deterministic regression or through parametric probabilistic policies. 
Such explicit action prediction can be insufficient in space-constrined parking spaces, where feasible maneuvers occupy only a narrow and highly structured region of the action space. 
As shown in Fig.~\ref{fig: intro}, even minor inaccuracies in trajectory generation may lead to collisions or excessive corrective maneuvers, ultimately reducing the overall success rate of parking \cite{jiang2025hope}.

To address these limitations, recent studies have explored diffusion policies, which formulate action prediction as a conditional denoising diffusion process \cite{chi2023diffusion}. 
By learning to recover clean actions from noisy inputs, diffusion policies provide a flexible generative framework for modeling complex and multi-modal action distributions, making them appealing for high-precision parking planning. Existing diffusion-based planners are commonly trained with imitation learning objectives, and recent studies further incorporate reinforcement learning (RL) mainly as a fine-tuning mechanism to improve task performance \cite{chi2023diffusion,wang2024diffusion,ren2024diffusion}. 
However, directly applying such diffusion policies to space-constrained parking remains nontrivial. 
Since feasible parking maneuvers occupy a small and highly structured region of the action space, generating high-precision actions from fully noisy initialization can be inefficient and unstable, especially when the planner must operate close to collision boundaries \cite{park2025flow}.

In this paper, we explore how to leverage diffusion models more effectively for reliable path planning, particularly with reinforcement learning (RL).
Rather than using RL merely as a fine-tuning mechanism for a diffusion policy, we use an RL-pretrained policy as an informative maneuver prior within the denoising process. 
Specifically, the action distribution predicted by the RL-pretrained policy is used to initialize the reverse denoising process at a truncation step, and subsequent denoising refines this coarse prior into more precise parking actions. 
However, under standard diffusion training, the intermediate distributions along the reverse process may not match the RL-induced prior, resulting in a distributional mismatch at the truncation point. 
To address this, we design a prior-aligned training procedure that aligns denoising trajectories with the RL-induced prior. 
This enables the model to exploit structured maneuvering knowledge during inference and improves planning reliability in precision-critical, space-constrained parking scenarios.

Overall, the contributions of this paper are:

\begin{itemize}
    \item To address the challenge of accurately modeling action distributions in constrained parking spaces, we propose \textbf{DRIP}, 
    a \textbf{D}iffusion-\textbf{R}ef\textbf{I}ned \textbf{P}lanner, which refines RL policies and enables RL-pretrained policies to be further optimized for generating precise and feasible maneuvers.
    \item To align denoising trajectories with prior action distributions, we design a denoising training procedure anchored in the priors, ensuring that the inference initialization remains consistent with the diffusion training process.
    \item We comprehensively evaluate DRIP in parking scenarios with varying levels of spatial constraints, demonstrating significant improvements in planning success and trajectory quality in complex environments.
\end{itemize}

\section{Related Work}
\subsection{Path Planning for Automated Parking}

Path planning in parking scenarios is challenging due to non-holonomic vehicle constraints, narrow operating spaces, and the need for non-smooth, multi-maneuver trajectories involving frequent gear shifts \cite{zheng2025multipark}. Traditional geometric and sampling-based approaches have been widely applied \cite{li2021optimization,dolgov2010path}, but their limited ability to handle complex environments restricts their effectiveness \cite{czubenko2015autonomous}.
With the advancement of deep learning, learning-based approaches have emerged as a promising alternative for parking path planning. By directly mapping environmental observations to action distributions, these methods alleviate the dependence on manually designed rules and cost functions. Existing studies can be broadly divided into imitation learning (IL) and reinforcement learning (RL) paradigms \cite{kiran2021deep}.

Imitation learning-based methods include approaches that imitate trajectories generated by rule-based planners \cite{chai2020design,chai2022deep}, methods that align simulated trajectories with human driving behaviors \cite{liu2017parking}, as well as behavior cloning techniques that directly learn input-output mappings from real vehicle data \cite{li2024parkinge2e}.
Reinforcement learning-based methods have also been widely investigated. For example, Du et al. trained agents in manually designed simulation environments \cite{du2020trajectory}, while Wang et al. reconstructed realistic environments from LiDAR data collected in the real world to support training \cite{wang2025rl}. Although directly applying standard RL algorithms to the planning problem is feasible for simple cases, studies that combine RL with heuristic search have demonstrated greater robustness \cite{bernhard2018experience}. To further enhance robustness across diverse parking scenarios, HOPE proposed a hybrid policy that integrates reinforcement learning with geometric constraints, explicitly enforcing kinematic feasibility and enabling reliable execution of complex multi-maneuver parking behaviors \cite{jiang2025hope}.

Despite their methodological differences, these approaches generally exhibit limited success rates in space-constrained parking scenarios. 
Both IL- and RL-based methods are typically formulated to learn a direct mapping from observations to actions. 
In practice, this is commonly realized through parameterized action outputs, including deterministic predictions or probabilistic policies such as unimodal or mixture Gaussian distributions. 
However, when the underlying action distribution is highly complex and accurate multi-step maneuvers are required, such explicit representations often struggle to faithfully capture the true set of feasible actions \cite{chi2023diffusion}. 
Due to inherent modeling inaccuracies arising from approximation errors, distributional variance, or imperfect value estimation \cite{zhang2019bridging,park2024value}, the resulting trajectories may deviate from feasible solutions, leading to collisions or excessive corrective maneuvers. This, in turn, degrades both planning efficiency and overall parking success rates.

\subsection{Diffusion Policy}
Diffusion models have recently demonstrated remarkable success across diverse domains \cite{ho2020denoising}. In particular, diffusion policies have emerged as a promising paradigm for action modeling by formulating policy learning as a conditional denoising process, which reconstructs structured trajectories from noisy inputs \cite{chi2023diffusion}. 
Building on this foundation, a variety of extensions have been introduced, such as enhanced spatial representations \cite{ze20243d}, accelerated inference schemes for real-time control \cite{dong2024diffuserlite}, guided diffusion processes for improved generation controllability \cite{jackson2024policy}, and self-supervised strategies that strengthen representation learning under data scarcity \cite{li2024crossway}. These developments underscore the flexibility of diffusion policies, yet they rely on imitation of training distributions without explicit mechanisms for exploration or reward-driven optimization, which constrains their effectiveness in long-horizon decision-making \cite{song2025survey}.

Given the strong sequential decision-making capability of reinforcement learning, recent research has explored integrating RL with diffusion \cite{zhu2023diffusion}. RL introduces reward-driven objectives that complement the generative strengths of diffusion models, enabling policies to go beyond imitation and actively optimize for higher returns \cite{janner2022planning}. 
One category of work achieves this through reject sampling, where Q-values are used to filter action candidates generated by a behavior cloning diffusion policy \cite{chen2022offline,hansen2023idql}. Another category employs weighted behavior cloning, reweighting diffusion-generated samples according to Q-values \cite{lu2023contrastive,ding2024diffusion}. While conceptually straightforward, these methods often suffer from low efficiency or remain limited in improving policy expressivity \cite{park2025flow}. An alternative approach is to integrate diffusion models into reparameterized policy gradient methods, treating the diffusion policy as the actor and fine-tune it directly with RL objectives \cite{ada2024diffusion}. Although this enables direct reward optimization, it relies on backpropagation through time (BPTT), which can introduce training instability and lead to suboptimal action choices \cite{ren2024diffusion}. In safety-critical tasks such as parking planning, these limitations can reduce the overall planning success rates.

In contrast to existing approaches, our setting leverages a pre-trained RL policy that already provides a reliable prior over actions. Rather than first pretraining a policy with IL and subsequently fine-tuning it with RL, we directly fine-tune the RL-pretrained policy through diffusion training, which could be in either an IL- or RL-based paradigm. This design not only refines the action distribution but also bypasses lengthy denoising procedures.

\section{Preliminaries}
\subsection{Problem Formulation}

The parking path planning is formulated as a sequential decision-making problem. At each discrete time step $t$, the vehicle pose is denoted by
$P_t = (x_t, y_t, \theta_t)$,
where $(x_t, y_t)$ is the rear-axle center position and $\theta_t$ is the heading angle.  
The state is denoted as
$s_t = \mathcal{S}(P_t)$,
which represents the environmental information obtained at pose $P_t$, including the target parking configuration, the occupancy map, and surrounding obstacles. Based on $s_t$, a policy $\pi_\theta$ generates an action
\begin{equation}
a_t \sim \pi_\theta(\cdot \mid s_t),
\end{equation}
where $a_t = (v_t, \delta_t)$ consists of the vehicle's velocity $v_t \in [v_{\min}, v_{\max}]$ and steering angle $\delta_t \in [\delta_{\min}, \delta_{\max}]$.

The vehicle dynamics follow the kinematic model \cite{althoff2017commonroad}, yielding a deterministic transition function $f:\mathbb{R}^3 \times \mathbb{R}^2 \to \mathbb{R}^3$:
\begin{align}
&P_{t+1} = f(P_t, a_t) = \nonumber\\
&\Bigl(
x_t + v_t \cos \theta_t \, \Delta t, 
 y_t + v_t \sin \theta_t \, \Delta t, 
 \theta_t + \tfrac{v_t}{L} \tan \delta_t \, \Delta t
\Bigr),
\end{align}
where $L$ is the wheelbase and $\Delta t$ is the discretization interval.

Overall, the parking path-planning task aims to learn a policy $\pi_\theta$ that generates a feasible trajectory from an initial vehicle pose $P_0$ to a target parking configuration $P_T$.  
A trajectory is represented as 
\begin{equation}
\tau = (P_0, P_1, \ldots, P_T),
\end{equation}
and evolves according to
\begin{equation}
P_{t+1} = f(P_t, a_t), \quad a_t \sim \pi_\theta(\cdot \mid s_t), \quad s_t = \mathcal{S}(P_t).
\end{equation}  
The terminal step $T$ is defined as the first time the vehicle reaches a state sufficiently close to the target configuration.  

The learned policy is therefore required to generate collision-free trajectories that respect vehicle kinematics and safety constraints. Such a policy can be obtained through different training paradigms, among which imitation learning and reinforcement learning are the most commonly adopted. 
\textit{Imitation learning} minimizes the discrepancy between the learned policy $\pi_\theta$ and an expert policy $\pi^*$, often through behavior cloning:
\begin{equation}
\min_\theta \; \mathbb{E}_{(s,a) \sim \mathcal{D}} \bigl[ \ell\big(\pi_\theta(a \mid s), \pi^*(a \mid s)\bigr) \bigr],
\end{equation}
where $\mathcal{D}$ is a dataset of expert demonstrations and $\ell(\cdot)$ is a loss function measuring the difference between the learned and expert actions.  
\textit{Reinforcement learning (RL)} optimizes the policy $\pi_\theta$ by maximizing the expected discounted return with discount factor $\gamma \in (0,1]$:
\begin{equation}
\theta^* 
= \arg\max_\theta \; \mathbb{E}_{\tau \sim \pi_\theta}\!\Big[ \sum_{t=0}^{T} \gamma^{\,t}\, r(s_t, a_t) \Big],
\end{equation}
where $r(s_t, a_t)$ denotes the reward signal obtained after executing action $a_t$ at state $s_t$.

\subsection{Diffusion Policy Preliminaries}

A diffusion policy models a conditional distribution over actions given the state by denoising from noise to a clean action. 
Let $a^{(0)}$ denote the ground-truth action paired with observation $o$ at state $s$. To avoid confusion, throughout this subsection we suppress the planning-time subscript (e.g., $a_t, s_t, o_t$) and simply write $a, s, o$.
The diffusion (denoising/noising) timestep is indexed by a superscript $k \in \{0,\ldots,K\}$, i.e., $a^{(k)}$ denotes the action at diffusion step $k$.

During training, a forward diffusion process progressively perturbs $a^{(0)}$ into a sequence $\{a^{(k)}\}_{k=1}^K$ under a variance schedule $\{\beta_k\}_{k=1}^K \subset (0,1)$ with
$\alpha_k := 1-\beta_k, 
\ 
\bar{\alpha}_k := \prod_{i=1}^{k}\alpha_i$ \cite{ho2020denoising}.
In practice, $\beta_k$ is typically chosen to increase with $k$ (e.g., linear or cosine schedules), so that $\alpha_k$ and $\bar{\alpha}_k$ \emph{decrease} monotonically as noise accumulates.

The step-wise forward (noising) kernel is
\begin{equation}
q\!\left(a^{(k)} \mid a^{(k-1)}\right)
= \mathcal{N}\!\bigl(\sqrt{\alpha_k}\,a^{(k-1)},\, \beta_k \mathbf{I}\bigr),
\quad k=1,\dots,K,
\end{equation}
which yields the marginal closed form
\begin{equation}
\label{eq: diffusion forward}
a^{(k)} = \sqrt{\bar{\alpha}_k}\,a^{(0)} + \sqrt{1-\bar{\alpha}_k}\,\epsilon,\;\; \epsilon \sim \mathcal{N}(0,\mathbf{I}).
\end{equation}

The reverse (denoising) process recovers $a^{(0)}$ from $a^{(K)} \sim \mathcal{N}(0,\mathbf{I})$ by iteratively reducing noise, using a neural network $\epsilon_\theta(a^{(k)},k,o)$ that predicts the noise in $a^{(k)}$ given the timestep $k$ and observation $o$ observed at state $s$. 
In the Denoising Diffusion Probabilistic Model (DDPM) \cite{ho2020denoising}, the reverse transition is Gaussian:
\begin{equation}
p_\theta\!\left(a^{(k-1)} \mid a^{(k)}, o\right)
= \mathcal{N}\!\Bigl(\mu_\theta\!\left(a^{(k)},k,o\right),\, \tilde{\beta}_k \mathbf{I}\Bigr),
\end{equation}
with mean
\begin{equation}
\mu_\theta\!\left(a^{(k)},k,o\right)
= \frac{1}{\sqrt{\alpha_k}}\!\left(
a^{(k)} - \frac{\beta_k}{\sqrt{1-\bar{\alpha}_k}}\, \epsilon_\theta(a^{(k)},k,o)
\right),
\end{equation}
and variance
\begin{equation}
\tilde{\beta}_k
= \frac{1-\bar{\alpha}_{k-1}}{1-\bar{\alpha}_{k}}\;\beta_k .
\end{equation}

Based on DDPM, the Denoising Diffusion Implicit Model (DDIM) \cite{song2020denoising} shares the same forward process but improves the reverse update that preserves desired marginals while allowing fewer denoising steps. 
Defining the predicted clean action
\begin{equation}
\label{eq: ddim 1}
\hat{a}^{(0)}(a^{(k)},k,o)
= \frac{a^{(k)} - \sqrt{1-\bar{\alpha}_k}\;\epsilon_\theta(a^{(k)},k,o)}{\sqrt{\bar{\alpha}_k}},
\end{equation}
the non-stochastic DDIM update can be written as: 
\begin{equation}
\label{eq: ddim 2}
\hat{a}^{(k-1)}
= \sqrt{\bar{\alpha}_{k-1}}\;\hat{a}^{(0)}(a^{(k)},k,o)
+ \sqrt{1-\bar{\alpha}_{k-1}}\;\epsilon_\theta(a^{(k)},k,o),
\end{equation}

Both DDPM and DDIM commonly adopt the same noise-prediction training objective by sampling $k \sim \mathcal{U}\{1,\dots,K\}$ and $\epsilon \sim \mathcal{N}(0,\mathbf{I})$:
\begin{equation}
\label{eq: dp loss}
\mathcal{L}_{\text{DP}}(\theta)
= \mathbb{E}_{(a^{(0)},o),\,k,\,\epsilon}
\Big[
\big\| \epsilon - \epsilon_\theta\big( a^{(k)},\; k,\; o\big) \big\|^2
\Big].
\end{equation}
In this paper, we adopt the DDIM method. At inference, the diffusion policy conditions on $s$ and executes the denoising process, initializing at $a^{(K)} \sim \mathcal{N}(0,\mathbf{I})$ and applying the update for $k=K,\dots,1$ to obtain $a^{(0)}$.

\section{Methodology}
\subsection{Diffusion-refinement Framework}
\label{sec: 4.A}

\begin{figure*}
  \centering
  \includegraphics[width=0.85\textwidth]{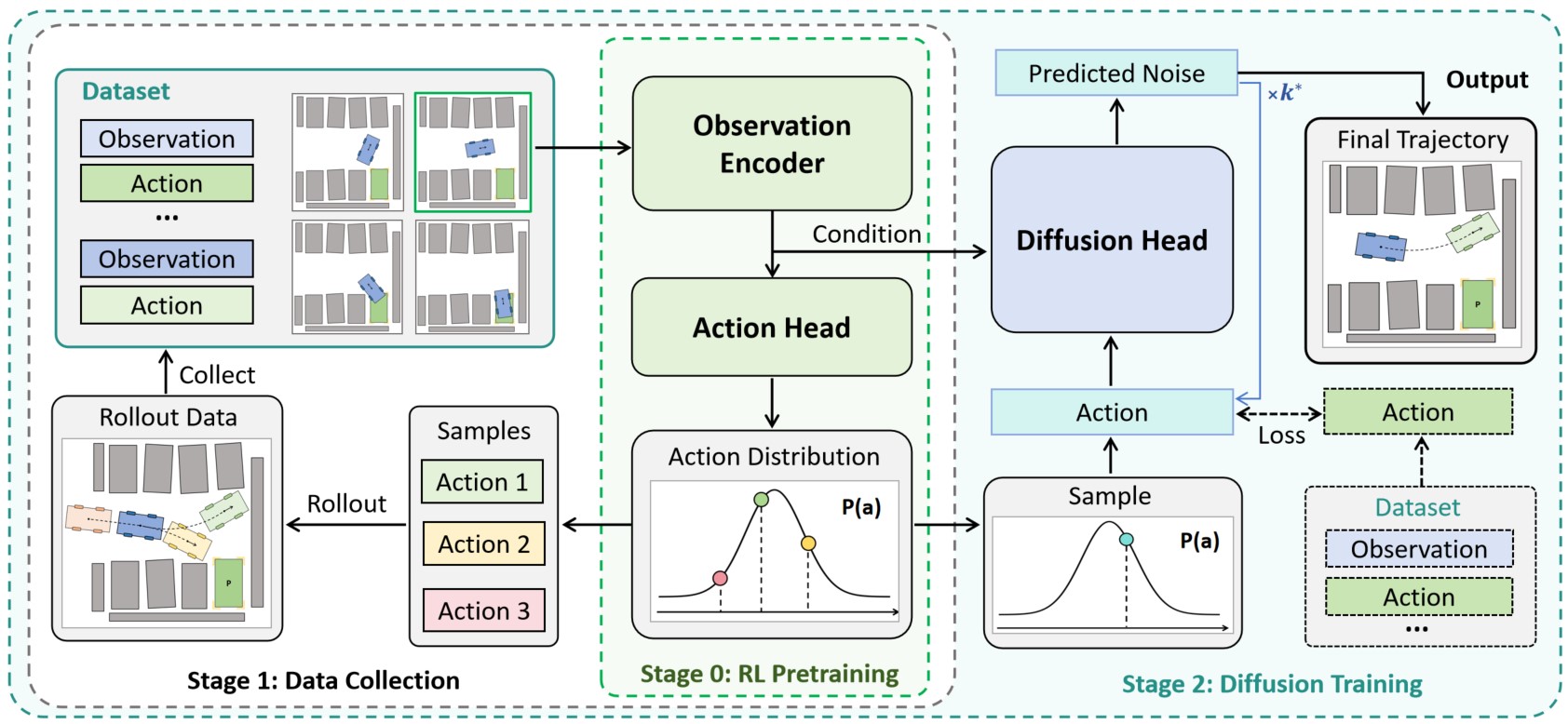}
  \caption{Overview of the proposed framework: (1) pretrain an RL policy to obtain a state-conditional prior distribution; (2) roll out this policy to collect a dataset; (3) train a diffusion policy on the collected dataset to refine the prior (via IL or RL).}
  \label{fig:overall_structure}
\end{figure*}

We adopt a diffusion-refined RL framework that starts from a pretrained policy $\pi_{\text{RL}}(a \mid o)$ mapping observations to actions. In practice, $\pi_{\text{RL}}$ is implemented with an observation encoder $f_{\phi}$ that extracts a latent representation $h = f_{\phi}(o)$, followed by a lightweight action head that outputs a parametric action distribution, typically a Gaussian $\mathcal{N}\!\big(\mu_{\text{RL}}(o), \sigma_{\text{RL}}^2(o)\big)$. While such parametric action modeling is effective in many settings, we observe that in space-constrained parking it can underfit the target action distribution. As a result, sampling from $\pi_{\text{RL}}(a \mid o)$ may occasionally yield suboptimal maneuvers that lead to collisions or trigger unnecessary corrective actions, degrading overall planning success.

To improve distributional accuracy without discarding existing RL knowledge, we adopt a self-distillation-style refinement pipeline. As shown in Fig. \ref{fig:overall_structure}, we first roll out the pretrained policy $\pi_{\text{RL}}(a|o)$ in the environment, sample actions from $\mathcal{N}\!\big(\mu_{\text{RL}}(o),\sigma_{\text{RL}}^2(o)\big)$ and record trajectories. Rollouts that reach the target position within a prescribed time step budget and remain collision-free are retained as successful demonstrations, and other trajectories are discarded. The resulting set of successful observation-action pairs serves as training data for diffusion refinement, allowing us to retain the strengths of $\pi_{\text{RL}}$ while filtering out its failure modes.

We then augment the pretrained architecture with a diffusion head to refine actions around the RL prior. The encoder $f_{\phi}$ is reused to provide conditioning $h = f_{\phi}(o)$ to a denoising network $\epsilon_{\theta}\!\big(a^{(k)},\,k,\,o\big)$, where $k \in \{0,\ldots,K\}$ indexes the diffusion (denoising/noising) step. Unlike standard diffusion policies that initialize from a standard Gaussian $a^{(K)} \sim \mathcal{N}(0, \mathbf{I})$, we initialize from the RL prior 
$a^{(k)} \sim \mathcal{N}\!\big(\mu_{\text{RL}}(o), \sigma_{\text{RL}}^2(o)\big)$ for some $k<K$
and perform a short reverse denoising chain to obtain a refined action $a^{(0)}$. 
During diffusion training, we sample pairs $(o, a)$ from the collected demonstration set and the corresponding action $a$ serves as the clean target $a^{(0)}$. Overall, this design leverages $\pi_{\text{RL}}$ as a structured prior and tasks the diffusion module with distributional refinement to sharpen and correct the RL outputs precisely in a constrained space.

\subsection{Prior-Aligned Diffusion}

\begin{figure}[t]
  \centering
  \includegraphics[width=\linewidth]{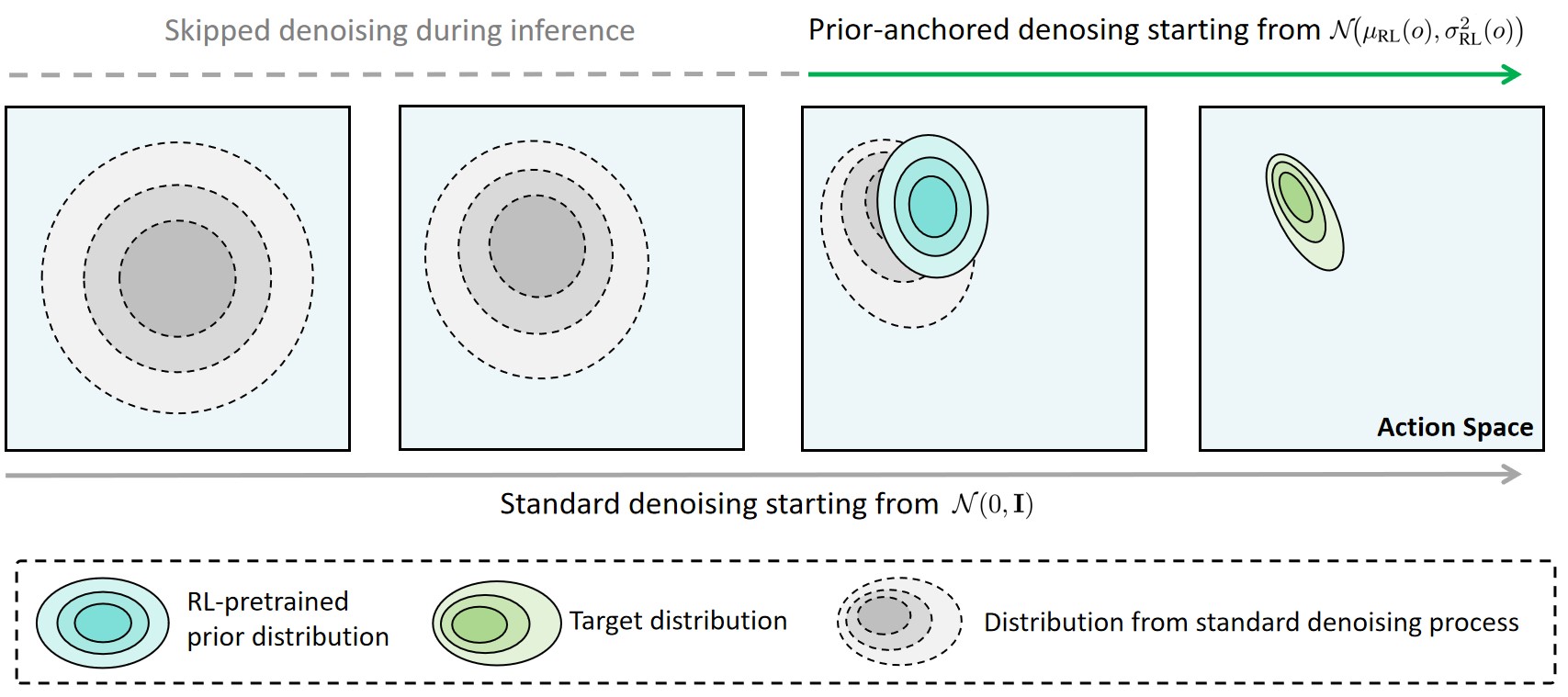}
  \caption{Illustration for the proposed prior-aligned diffusion training. Standard diffusion starts from random noise and yields an intermediate distribution that does not align with the RL-pretrained prior distribution. Our goal is to match the distribution from the denoising process with the prior distribution through alignment training, so the denoising process can start from the prior distribution at inference. }
  \label{fig:distribution}
  \vspace{5pt}
\end{figure}

When deploying the planner, we initialize the reverse (denoising) process from the pretrained RL prior and execute only the last $k$ denoising steps for some truncation index $k$. However, under standard diffusion training, the marginal at an intermediate step $k$ follows the forward relation in Eq.~\eqref{eq: diffusion forward}, whose mean depends on $a^{(0)}$ and whose standard deviation equals $\sqrt{1-\bar{\alpha}_{k}}$. As shown in Fig.~\ref{fig:distribution}, this does \emph{not} match the RL prior:
\begin{equation}
\label{eq: RL distribution}
p_{RL}(a \mid o) \;=\; \mathcal{N}\!\big(\mu_{\text{RL}}(o),\, \sigma_{\text{RL}}^{2}(o)\big),
\end{equation}
which creates a distributional mismatch at the truncation point and induces biased denoising updates, ultimately degrading action accuracy. To ensure training-inference consistency, we therefore modify the noising procedure during training so that the reverse trajectory is \emph{prior-aligned}. For clarity, we omit the observation $o$ in the derivations below (all quantities are conditioned on $o$ unless stated otherwise).

\subsubsection{Introduction of Time-Variant Mean}
We desire the truncated initialization $a^{(k^*)}$ to match the RL prior in Eq.~\eqref{eq: RL distribution} for some $k^*<K$, so that inference can start at $k=k^*$ rather than from pure noise. To this end, we replace the standard forward parameterization (as shown in Eq.~\eqref{eq: diffusion forward}) with a prior-centered mean schedule:
\begin{equation}
\label{eq: modified forward}
a^{(k)} \;=\; \sqrt{\bar{\alpha}_{k}}\,\mu(k) \;+\; \sqrt{1-\bar{\alpha}_{k}}\,\epsilon, 
\qquad \epsilon \sim \mathcal{N}(0,\mathbf{I}),
\end{equation}
where $\mu(k)$ is a $k$-dependent mean to be designed. The truncation index $k^{*}$ is selected to match the standard deviation of $a^{(k)}$ to that of the RL prior:
\begin{equation}
\label{eq: k selection}
k^{*} \;=\; \arg\min_{k}\; \big|\, \sigma_{\text{RL}} \;-\; \sqrt{\,1-\bar{\alpha}_{k}\,}\,\big| .
\end{equation}
Training-inference consistency further requires that the mean schedule reproduces the correct marginals at the endpoints:
\begin{equation}
\label{eq: mean bcs}
\begin{aligned}
\mathbb{E}\!\left[a^{(0)}\right]   &= \sqrt{\bar{\alpha}_{0}}\,\mu(0)   = a^{(0)}, \\
\mathbb{E}\!\left[a^{(k^*)}\right] &= \sqrt{\bar{\alpha}_{k^*}}\,\mu(k^*) = \sqrt{\bar{\alpha}_{k^*}}\,\mu_{\text{RL}} \, .
\end{aligned}
\end{equation}
We parameterize the schedule as a two-term interpolation,
\begin{equation}
\label{eq: mu blend}
\mu(k) \;=\; w_1(k)\,\mu_{\text{RL}} \;+\; w_2(k)\,a^{(0)},
\end{equation}
and constrain the weights to satisfy Eq.~\eqref{eq: mean bcs}; in particular,
\begin{equation}
    \label{eq: coefficients}
    w_1(k^*)= w_2(0) = 1,\quad w_1(0)= w_2(k^*) = 0.
\end{equation}
A concrete schedule for $(w_1,w_2)$ will be specified in Sec.~\emph{Selection of Coefficients}, where we additionally ensure numerically stable scaling of the training loss near small $k$.

\subsubsection{Training Objective}

With the time-variant mean $\mu(k)$, the forward (noising) process in Eq.~\eqref{eq: modified forward} can be rewritten as
\begin{equation}
\begin{aligned}
&a^{(k)} 
= \sqrt{\bar{\alpha}_{k}}\,\mu(k) \;+\; \sqrt{1-\bar{\alpha}_{k}}\,\epsilon \\[2pt]
&= \sqrt{\bar{\alpha}_{k}}\,a^{(0)} 
+ \sqrt{1-\bar{\alpha}_{k}}\left(
\sqrt{\tfrac{\bar{\alpha}_{k}}{1-\bar{\alpha}_{k}}}[\mu(k)-a^{(0)}] + \epsilon
\right).
\end{aligned}
\end{equation}
Comparing with Eq.~\eqref{eq: diffusion forward}, the corresponding noise-prediction objective becomes
\begin{equation}
\label{eq: new loss}
\mathcal{L}_{\text{DP}}(\theta)=\mathbb{E}_{a^{(0)},k,\epsilon}\Big[\big\|\epsilon+\sqrt{\tfrac{\bar{\alpha}_{k}}{1-\bar{\alpha}_{k}}}\,(\mu(k)-a^{(0)})-\epsilon_{\theta}(a^{(k)},k)\big\|^{2}\Big].
\end{equation}
This indicates that during training the Gaussian perturbation $\epsilon \sim \mathcal{N}(0,\mathbf{I})$ is applied around the scheduled mean $\mu(k)$ rather than the clean action $a^{(0)}$.

\subsubsection{Selection of Coefficients}
A direct concern in Eq.~\eqref{eq: new loss} is the factor $\sqrt{\bar{\alpha}_{k}/(1-\bar{\alpha}_{k})}$, which can become large as $k\!\to\!0$. To stabilize learning, we choose $(w_1,w_2)$ so that the terms coupled with $1/\sqrt{1-\bar{\alpha}_{k}}$ remain bounded. Using Eqs.~\eqref{eq: mu blend} and \eqref{eq: mean bcs},
\begin{equation}
\begin{aligned}
\label{eq: potential unbounded}
&\sqrt{\tfrac{\bar{\alpha}_{k}}{\,1-\bar{\alpha}_{k}\,}}\,[\,\mu(k)-a^{(0)}\,]
\;= \\
&\quad\ w_1(k)\,\sqrt{\tfrac{\bar{\alpha}_{k}}{\,1-\bar{\alpha}_{k}\,}}\;\mu_{\text{RL}}
\;-\;
\big(1-w_2(k)\big)\,\sqrt{\tfrac{\bar{\alpha}_{k}}{\,1-\bar{\alpha}_{k}\,}}\;a^{(0)}.
\end{aligned}
\end{equation}

Since $\mu_{\text{RL}}$, $a^{(0)}$, and $\sqrt{\bar{\alpha}_{k}}$ are bounded, the only potentially unbounded factors in Eq.~\eqref{eq: potential unbounded} come from the ratios $w_1(k)/\sqrt{1-\bar{\alpha}_{k}}$ and $\big(w_2(k)-1\big)/\sqrt{1-\bar{\alpha}_{k}}$. 
Therefore, it suffices to impose the bounds
\begin{equation}
\label{eq: upper bound}
\left|\,\frac{w_1(k)}{\sqrt{\,1-\bar{\alpha}_{k}\,}}\,\right| \le U,
\qquad
\left|\,\frac{1-w_2(k)}{\sqrt{\,1-\bar{\alpha}_{k}\,}}\,\right| \le U,
\end{equation}
for some constant $U>0$, together with the endpoint constraints in Eq.~\eqref{eq: mean bcs}.

To meet these requirements, we choose a schedule that interpolates with the accumulated noise standard deviation:
\begin{equation}
\label{eq: coeff schedule}
\begin{aligned}
w_1(k) &=
\begin{cases}
\sqrt{\dfrac{1-\bar{\alpha}_{k}}{\,1-\bar{\alpha}_{k^{*}}\,}}, & k \le k^{*},\\[6pt]
1, & k > k^{*},
\end{cases}\\[8pt]
w_2(k) &=
\begin{cases}
1-\sqrt{\dfrac{1-\bar{\alpha}_{k}}{\,1-\bar{\alpha}_{k^{*}}\,}}, & k \le k^{*},\\[6pt]
0, & k > k^{*}.
\end{cases}
\end{aligned}
\end{equation}
This schedule satisfies the requirements in Eqs.~\eqref{eq: mean bcs} and \eqref{eq: coefficients}. Moreover, Eq.~\eqref{eq: upper bound} holds with $U = 1/\sqrt{\,1-\bar{\alpha}_{k^{*}}\,}$. 
In practice, $U$ remains moderate. For example, under a typical cosine noise schedule and $k^{*}=K/5$, we have  $U  \approx 3$.
The resulting $(w_1,w_2)$ are monotone in $k$, yield a smooth interpolation from $a^{(0)}$ to $\mu_{\text{RL}}$, and guarantee numerically stable scaling of the training objective near $k\!\to\!0$.

\subsection{Training-Inference Pipeline and Learning Objectives}

Using Eqs.~\eqref{eq: mu blend} and \eqref{eq: coeff schedule}, the time-variant mean $\mu(k)$ is fully specified as
\begin{equation}
\label{eq: mu final}
\begin{aligned}
    \mu(k) = \begin{cases}
       \sqrt{\frac{1-\bar{\alpha}_k}{1-\bar{\alpha}_{k^*}}}\,\mu_{RL}  +  \left( 1-\sqrt{\frac{1-\bar{\alpha}_k}{1-\bar{\alpha}_{k^*}}}\,  \right) a^{(0)} \, , \  k \le k^* \\
       \mu_{RL}, \qquad \qquad  k > k^*
    \end{cases}
\end{aligned}
\end{equation}
and the prior-aligned forward process in Eq.~\eqref{eq: modified forward} yields the diffusion imitation-learning loss in Eq.~\eqref{eq: new loss}.

\begin{algorithm}[bth]
\caption{Training Procedure}
\label{alg: train}
\begin{algorithmic}[1]
\REQUIRE Pretrained $\pi_{\mathrm{RL}}(a\mid o)$, dataset $\mathcal{D}$ from $\pi_{\mathrm{RL}}$ rollouts; schedule $\{\beta_{k}\}_{k=1}^{K}$ and $\{\bar{\alpha}_{k}\}_{k=1}^{K}$; learning rate $\eta$
\STATE Initialize diffusion head $\epsilon_\theta$
\STATE Initialize critic network $\tilde{Q}_{\psi}$ (Only offline RL)
\STATE Calculate truncation step $k^{*}$ via Eq.~\eqref{eq: k selection}
\FOR{Iteration $i = 1$ \TO $N$}
  \STATE Sample $(o, a^{(0)}) \sim \mathcal{D}$, $k \sim \mathrm{Unif}\{1,\ldots,K\}$
  \STATE Form RL prior $\mathcal{N}\!\big(\mu_{\text{RL}}(o),\,\sigma_{\text{RL}}^{2}(o)\big)$ with $\pi_{\mathrm{RL}}(a\mid o)$
  \STATE Build $\mu(k)$ using Eq.~\eqref{eq: mu final}
  \STATE Form $a^{(k)}$ by adding noise to $\mu(k)$ using Eq.~\eqref{eq: modified forward}
  \STATE Predict the noise $ \epsilon_\theta(a^{(k)}, k, o)$
  \STATE Compute $\mathcal{L}_{\text{DP}}(\theta)$ by Eq.~\eqref{eq: new loss}
  \IF{Offline RL}
     \FOR{$k = k^{*}$ \TO $1$}
        \STATE Predict action $\hat{a}^{(k-1)}$ using Eqs.~\eqref{eq: ddim 1} and \eqref{eq: ddim 2}
     \ENDFOR
     \STATE Set $\mathcal{L}(\theta) \leftarrow \mathcal{L}_{\text{DP}}(\theta) \;-\; \lambda\,\tilde{Q}_\psi\!\big(o, \hat{a}^{(0)}\big)$ \quad (cf. Eq.~\eqref{eq: dql})
  \ELSE
     \STATE Set $\mathcal{L}(\theta) \leftarrow \mathcal{L}_{\text{DP}}(\theta)$
  \ENDIF
  \STATE Update $\theta \leftarrow \theta - \eta\,\nabla_\theta \mathcal{L}(\theta)$
\ENDFOR
\end{algorithmic}
\end{algorithm}

\begin{algorithm}[t]
\caption{Inference Procedure}
\label{alg: Inference}
\begin{algorithmic}[1]
\REQUIRE Observation $o$, truncation index $k^{*}$
\STATE Form RL prior $\mathcal{N}\!\big(\mu_{\text{RL}}(o),\,\sigma_{\text{RL}}^{2}(o)\big)$ with $\pi_{\mathrm{RL}}(a\mid o)$
\STATE Sample $a^{(k^{*})} \sim \mathcal{N}\!\big(\mu_{\text{RL}}(o),\,\sigma_{\text{RL}}^{2}(o)\big)$
\FOR{$k = k^{*}$ \TO $1$}
  \STATE Predict the noise $ \epsilon_\theta(a^{(k)}, k, o)$
  \STATE Obtain action $\hat{a}^{(k-1)}$ using Eqs.~\eqref{eq: ddim 1} and \eqref{eq: ddim 2}
\ENDFOR
\STATE Output $\hat{a}^{(0)}$
\end{algorithmic}
\end{algorithm}

Note that our use of RL at the first stage is purely pretraining (to provide a structured prior and rollouts). This usage is orthogonal to prior work that optimizes diffusion with RL, which means our method can be seamlessly augmented with existing offline diffusion RL training approaches. We retain $\mathcal{L}_{\text{DP}}$ as a BC regularizer and add a value-maximization term evaluated on the final action produced by the truncated DDIM reverse chain shown in Eqs.~\eqref{eq: ddim 1} and \eqref{eq: ddim 2}. 
Concretely, starting from $a^{(k^*)}\!\sim\!\mathcal{N}\!\big(\mu_{\text{RL}},\sigma_{\text{RL}}^2\big)$, we apply the deterministic DDIM updates for $k=k^*,\ldots,1$ and denote the final action by $\hat{a}^{(0)}$. 
Let $\tilde{Q}_\psi(o,a)$ denote a value estimator that could either be a single critic $Q_\psi(o,a)$ or the minimum over twin critics $\text{min}\{Q_{\psi_1}(o,a), Q_{\psi_2}(o,a)\}$ in practice, the diffusion-offline RL objective (in the spirit of diffusion Q-learning \cite{wang2022diffusion}) is
\begin{equation}
\label{eq: dql}
\mathcal{L}_{\text{DQL}}(\theta)
=
\mathcal{L}_{\text{DP}}(\theta)
-
\lambda\,\mathbb{E}_{o,a^{(k^*)}\sim \mathcal{N}(\mu_{\text{RL}},\sigma_{\text{RL}}^2)}\!\Big[
\tilde{Q}_\psi\big(o,\hat{a}^{(0)}\big)
\Big],
\end{equation}
where $\lambda>0$ is a coefficient to balance BC regularization and value maximization.

Overall, the training procedure is summarized in Alg.~\ref{alg: train}, where the forward process and the objective are modified to align diffusion training with the RL prior. At inference, as shown in Alg.~\ref{alg: Inference}, we initialize from the RL prior and run a truncated denoising chain to obtain the final action $\hat{a}^{(0)}$.

\section{Experiment}
\subsection{Experimental Setup}

We evaluate our method on the HOPE benchmark~\cite{jiang2025hope}, which composes diverse parking scenes from the open-source DLP dataset and a 2D simulator~\cite{shen2022parkpredict+,li2023tactics2d}. 
The simulated vehicle is configured as an SUV-like passenger car with a length of 4.52~m, a width of 1.86~m, and a minimum turning radius of 6.27~m. 
Scenarios are categorized into three difficulty levels, \emph{normal}, \emph{medium}, and \emph{hard}, according to spatial narrowness following ISO~20900~\cite{ISO}. 
The training set includes $2{,}000$ hard-level scenes, $3{,}000$ medium-level scenes, and $2{,}000$ normal-level scenes. 
In total, we use $7{,}000$ scenes for training, $700$ for validation, and another $700$ for testing. 
From the pretrained RL policy rollouts discussed in Sec.~\ref{sec: 4.A}, we collect $158{,}706$ successful state--action pairs from the $7{,}000$ training scenes for diffusion training. 
For evaluation, each testing scene is repeated 10 times, and the final success rate is computed by averaging over repeated trials for each difficulty level and over all test scenarios.

At evaluation time, each episode is limited to $50$ interaction steps. Failing to reach the target configuration within this budget is counted as a planning failure. 
For fairness, all methods enable \emph{analytic expansions} \cite{dolgov2010path}, a shortcut practice commonly used in unstructured planning.
We report planning success rate on the $700$-scene test set and also break down results by difficulty.

Following the benchmark, the observation at each step comprises: (i) an ego-centered bird’s-eye-view (BEV) occupancy raster in which occupied cells (obstacles and parked vehicles) are 1 and drivable are 0;  (ii) radial obstacle distances sampled every $10^\circ$, and (iii) the relative parking-slot configuration in the ego frame (position, heading, and slot orientation). 
The action space is two-dimensional, and each step outputs a path increment (step length) and a steering command.

\begin{table}[th]
\centering
\caption{Key hyperparameters.}
\label{tab: hyperparams}
\begin{tabular}{ll}
\toprule
Hyperparameter & Value \\
\midrule
RL pretrain episodes & $100{,}000$ \\
Diffusion epochs & 60 \\
Training batch size & 64 \\
Optimizer  & AdamW \\
Diffusion learning rate & $1\times 10^{-4}$ \\
Encoder learning rate & $1\times 10^{-5}$ \\
DDIM train steps & 100 \\
DDIM inference steps & 25 \\
Truncation step $k^*$ & 5 \\
DDIM $\beta_k$ range & $[0.0001,\ 0.02]$ \\
Noise schedule  & Cosine \\
RL discount $\gamma$ & 0.98 \\
DQL weight $\lambda$ & 1 \\
\bottomrule
\end{tabular}
\vspace{0pt}
\end{table}

All learning-based baselines and our approach share the same observation encoder, a pretrained transformer-based module that produces a $512$-D representation \cite{jiang2025hope}. We then use this feature as the diffusion condition, followed by either an action head or a diffusion head that outputs the $2$-D action.

Training hyperparameters are summarized in Table~\ref{tab: hyperparams}.
All diffusion-based methods use the same $\{\beta_k\}_{k=1}^{K}$ schedule and accumulated products $\{\bar{\alpha}_k\}_{k=1}^{K}$, and the truncation index $k^\ast$ is selected by Eq.~\eqref{eq: k selection}. Experiments are carried out on a single NVIDIA RTX-3090 and a 32-core AMD EPYC 7542 CPU.

\subsection{Experiment Results}

\renewcommand{\arraystretch}{1.15}
\begin{table}[b]
\centering
\caption{Success rate (\%) under different difficulties.}
\label{tab: results}
\small
\begin{tabular}{c|c|ccc|c}
\hline
\textbf{Paradigm} & \textbf{Method} & \textbf{Hard} & \textbf{Medium} & \textbf{Normal} & \textbf{Avg.} \\
\hline
 / & HA* & 6.0 & 74.3 & 82.5 & 57.1 \\
\hline
\multirow{2}{*}{\makecell{Pretrain\\ (RL)}} 
 & PPO & 28.8 & 81.6 & 83.8 & 67.1 \\& SAC  
 & 43.2 & 85.5 & 92.9 & 75.5 \\
\hline
\multirow{3}{*}{\makecell{Diffusion\\ (IL)}}
 & DP & 48.3 & 85.2 & 92.7 & 76.8 \\
 &  + RL init. & 46.6 & 88.5 & 91.4 & 77.4  \\
 & DRIP (ours) & \textbf{59.1} & \textbf{88.9} & \textbf{96.2} & \textbf{82.4} \\
\hline
\multirow{3}{*}{\makecell{Diffusion\\ (RL)}}
 & DQL & 45.0 & 86.4 & 91.0 & 75.9 \\
 &  + RL init. & 51.0 & 89.0 & 94.4 & 79.7  \\
 & DRIP (ours) & \textbf{65.9} & \textbf{92.0} & \textbf{96.8} & \textbf{85.9} \\
\hline
\end{tabular}
\end{table}

\newcommand{\deltag}[1]{\textcolor{green!60!black}{\scriptsize\, (#1)}}
\newcommand{\deltar}[1]{\textcolor{red!70!black}{\scriptsize\, (#1)}}
\begin{table*}[b]
\caption{Comparison of metrics for trajectory quality}
\centering
\small
\setlength{\tabcolsep}{5.5pt}
\renewcommand{\arraystretch}{1.12}
\begin{tabular}{l|l|cc|cc|cc}
\toprule
Difficulty & Method &
\multicolumn{2}{c|}{Path length (m) $\downarrow$} &
\multicolumn{2}{c|}{Interaction steps $\downarrow$} &
\multicolumn{2}{c}{Gear shifts $\downarrow$} \\
\cline{3-8}
 &  & Avg & Avg-Best & Avg & Avg-Best & Avg & Avg-Best \\
\midrule

\multirow{2}{*}{Normal} &
RL   & 18.68 & 15.34 & 20.45 & 15.38 &  7.03 &  3.50 \\
& DRIP & 16.68\deltag{-2.00} & 15.35\deltar{+0.01} &
        17.91\deltag{-2.54} & 15.58\deltar{+0.20} &
         5.17\deltag{-1.86} &  3.69\deltar{+0.19} \\
\midrule

\multirow{2}{*}{Medium} &
RL   & 22.55 & 17.98 & 24.90 & 18.45 &  8.52 &  3.53 \\
& DRIP & 19.52\deltag{-3.03} & 17.71\deltag{-0.27} &
        21.44\deltag{-3.46} & 18.29\deltag{-0.16} &
         5.59\deltag{-2.93} &  3.65\deltar{+0.12} \\
\midrule

\multirow{2}{*}{Hard} &
RL   & 30.68 & 20.90 & 40.29 & 30.22 & 22.82 & 12.64 \\
& DRIP & 25.56\deltag{-5.12} & 18.99\deltag{-1.91} &
        33.39\deltag{-6.90} & 26.02\deltag{-4.20} &
        17.00\deltag{-5.82} & 10.51\deltag{-2.13} \\
\midrule

\multirow{2}{*}{Overall} &
RL   & 23.77 & 18.06 & 28.02 & 20.94 & 12.18 &  6.12 \\
& DRIP & 20.44\deltag{-3.33} & 17.40\deltag{-0.66} &
        23.84\deltag{-4.18} & 19.72\deltag{-1.22} &
         8.73\deltag{-3.45} &  5.62\deltag{-0.50} \\
\bottomrule
\end{tabular}
\label{tab:drip_vs_rl_with_delta}
\vspace{-10pt}
\end{table*}

We compare against a widely used non-learning planning method and planners trained with learning-based methods. 

\begin{itemize}
\item \textbf{Hybrid A* (HA*)} \cite{dolgov2010path}: A non-learning heuristic-search-based planner widely used for unstructured planning.

\item \textbf{Reinforcement Learning (RL):} Following the HOPE training protocol \cite{jiang2025hope}, we apply two planners, an on-policy \textbf{PPO}-based planner and an off-policy \textbf{SAC}-based planner. These serve both as competitive baselines and as the source of pretrained policies used for parameter initialization in diffusion-based methods.

\item \textbf{Diffusion Policy (DP)} \cite{chi2023diffusion}: A diffusion planner that models the policy via conditional denoising and is trained with imitation learning. We apply the version (\textbf{DP}), which is trained from scratch using the IL objective (Eq.~\eqref{eq: dp loss}) as applied in the original paper, and the variant (\textbf{ + RL init.}), which shares the same training process but initializes the backbone using the RL-pretrained parameters.

\item \textbf{Diffusion Q-Learning (DQL)} \cite{wang2022diffusion}: A diffusion actor optimized with an offline RL objective. We also report two versions of this method. The approach (\textbf{DQL}), as applied in the original paper, uses the Q-learning loss with behavior cloning regularization (Eq.~\eqref{eq: dql}) without RL pretraining, and the version (\textbf{+ RL init.}) initializes the backbone with the RL-pretrained parameters.

\item \textbf{Diffusion Refined Planner (DRIP)}. The proposed method that (i) starts the reverse process from the pretrained RL-pretrained prior distribution at a truncation step, and (ii) aligns the forward process to that prior via the proposed prior-alignment training process. We evaluate our methods trained under both the imitation-learning paradigm (IL) and the reinforcement learning paradigm (RL). In both cases, the backbone is initialized from the pretrained policy.
\end{itemize}

\subsubsection{Planning success rate}
Table~\ref{tab: results} reports success rates for the above methods. 
Hybrid A* exhibits low success on hard cases and modest performance overall, whereas RL-pretrained agents attain better success across difficulty levels, indicating the benefit of data-driven learning. 
Similarly, DP and DQL achieve performance comparable to RL baselines, but performance gains are limited when trained from scratch. This indicates that learning to denoise from pure noise without an informative prior is challenging in a constrained space configuration. 
Initializing from an RL-pretrained policy improves the diffusion baselines slightly by \textbf{+0.6\%} for DP and \textbf{+3.8\%} for DQL, with the larger gain for DQL plausibly due to better alignment between the RL-style objective used in both pretraining and the second-stage diffusion training.

In contrast, our Diffusion Refinement approach yields consistent and substantial improvements. 
Relative to their RL-initialized counterparts, \textbf{DRIP} improves overall success by \textbf{+5.0\%} and \textbf{+6.2\%}, respectively. Compared to training from scratch, the gains are \textbf{+5.6\%} (over DP) and \textbf{+10.0\%} (over DQL). 
Improvements are most pronounced on \emph{hard} scenes, with a gain of \textbf{+10.8\%} over DP and \textbf{+10.9\%} over DQL. 
These results support the effectiveness of our proposed diffusion-based refinement method, not mere parameter initialization, for learning a better policy in space-constrained parking.

\newlength{\visfigheight}
\setlength{\visfigheight}{4.2cm}

\begin{figure*}[htb]
\centering
\captionsetup[subfigure]{labelformat=empty}

\subfloat[(a-1)]{%
  \begin{minipage}[t]{0.24\textwidth}\centering
    \includegraphics[height=\visfigheight]{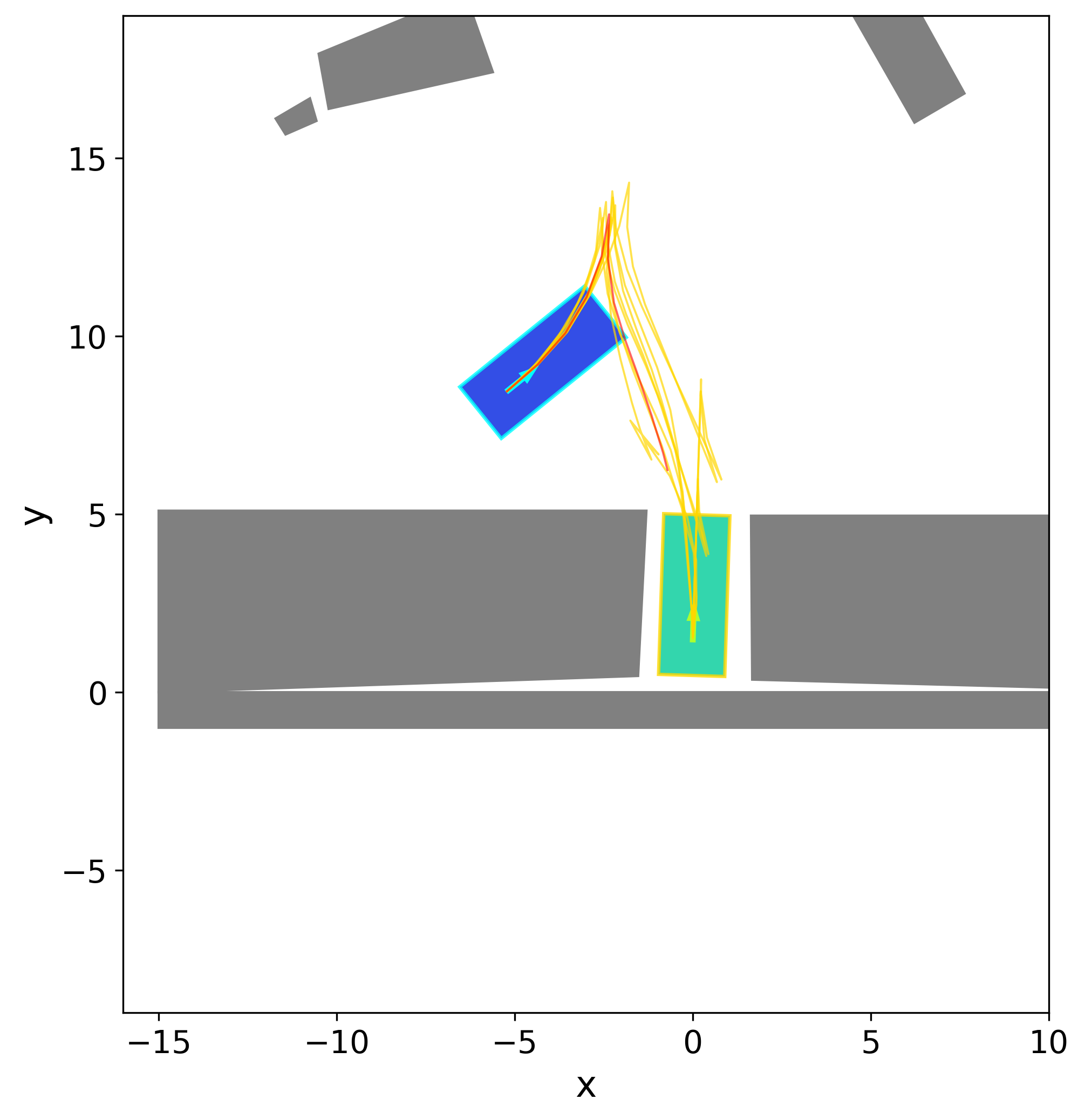}
  \end{minipage}}
\hfill
\subfloat[(b-1)]{%
  \begin{minipage}[t]{0.24\textwidth}\centering
    \includegraphics[height=\visfigheight]{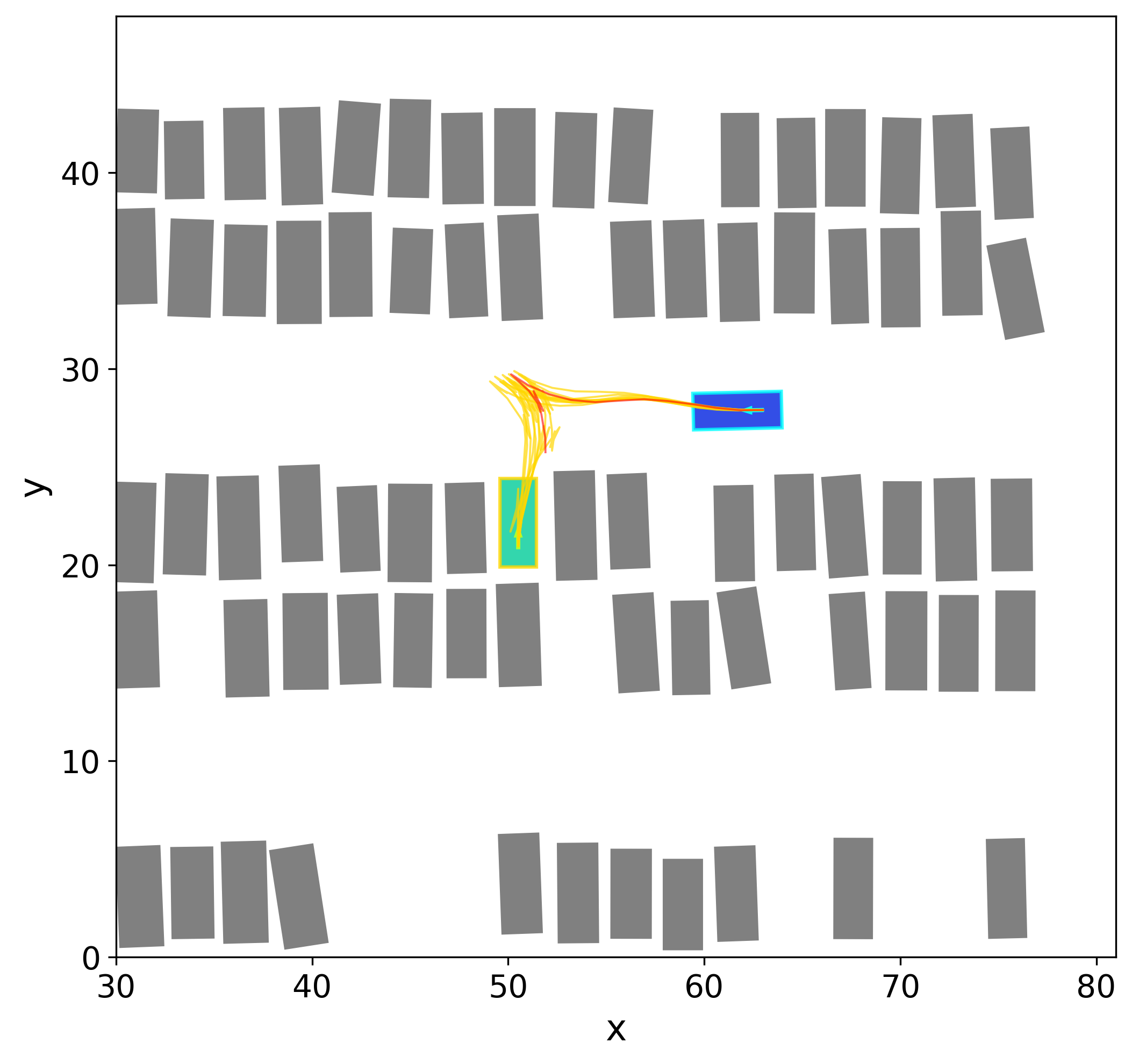}
  \end{minipage}}
\hfill
\subfloat[(c-1)]{%
  \begin{minipage}[t]{0.24\textwidth}\centering
    \includegraphics[height=\visfigheight]{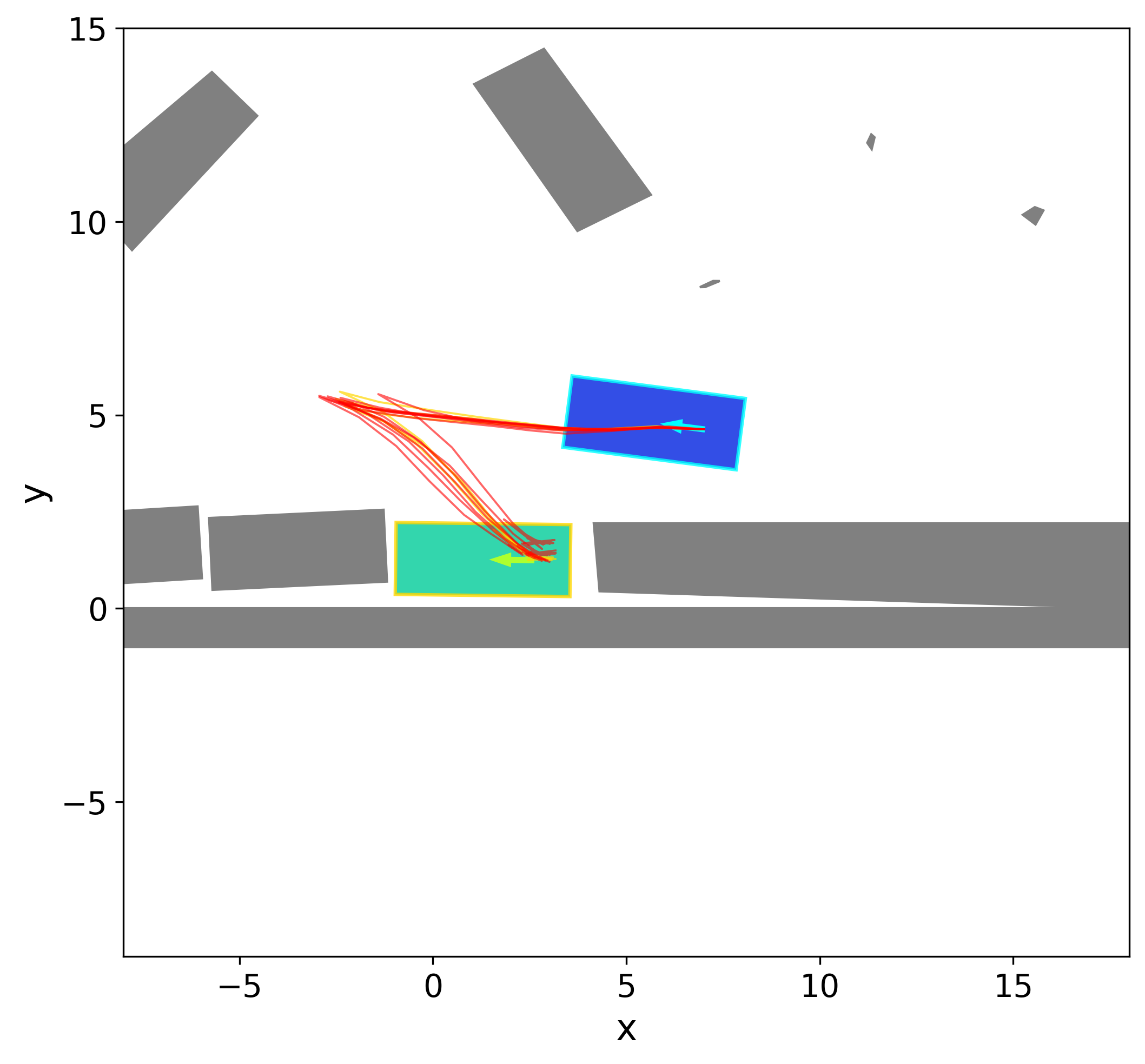}
  \end{minipage}}
\hfill
\subfloat[(d-1)]{%
  \begin{minipage}[t]{0.24\textwidth}\centering
    \includegraphics[height=\visfigheight]{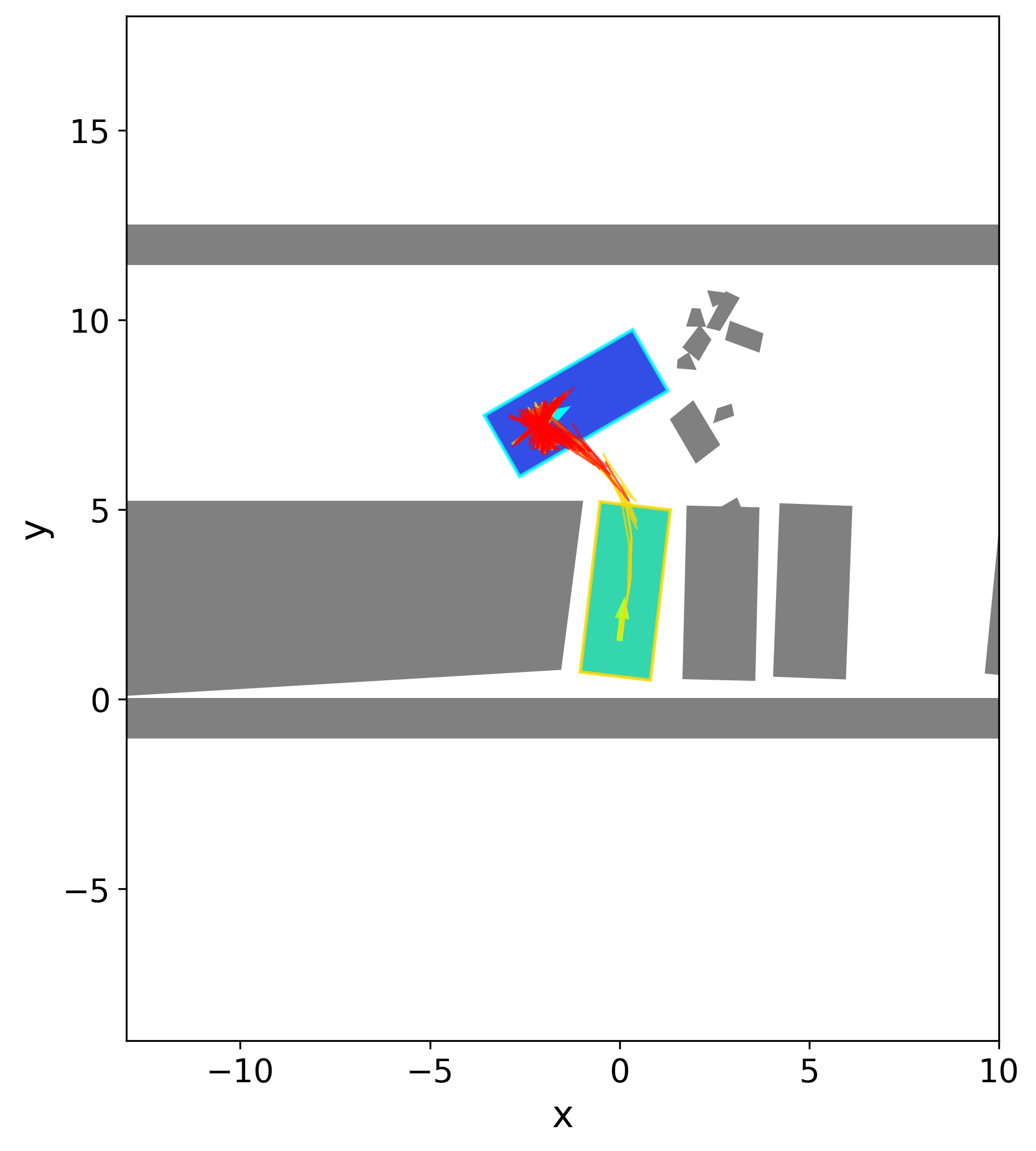}
  \end{minipage}}


\subfloat[(a-2)]{%
  \begin{minipage}[t]{0.24\textwidth}\centering
    \includegraphics[height=\visfigheight]{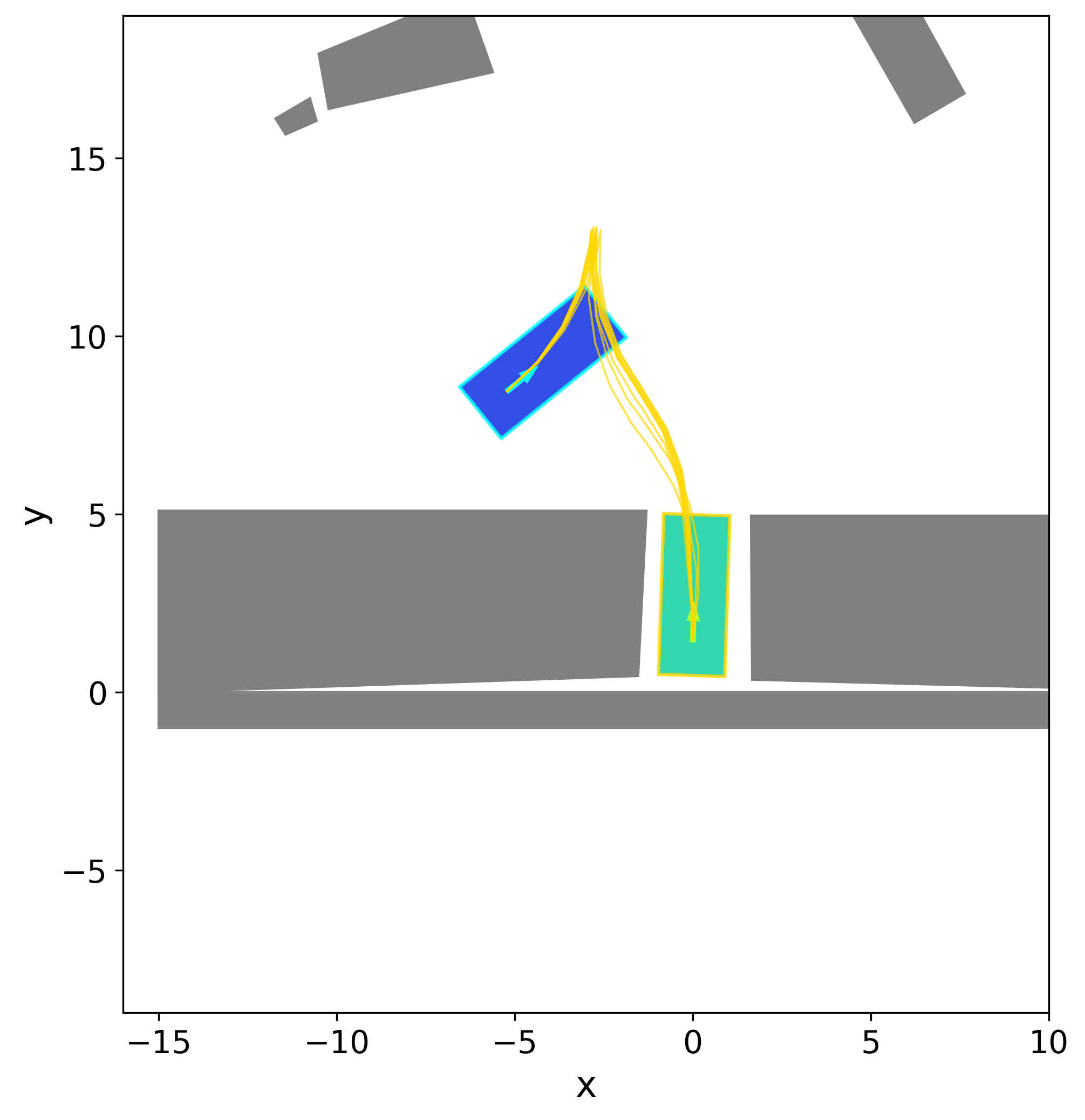}
  \end{minipage}}
\hfill
\subfloat[(b-2)]{%
  \begin{minipage}[t]{0.24\textwidth}\centering
    \includegraphics[height=\visfigheight]{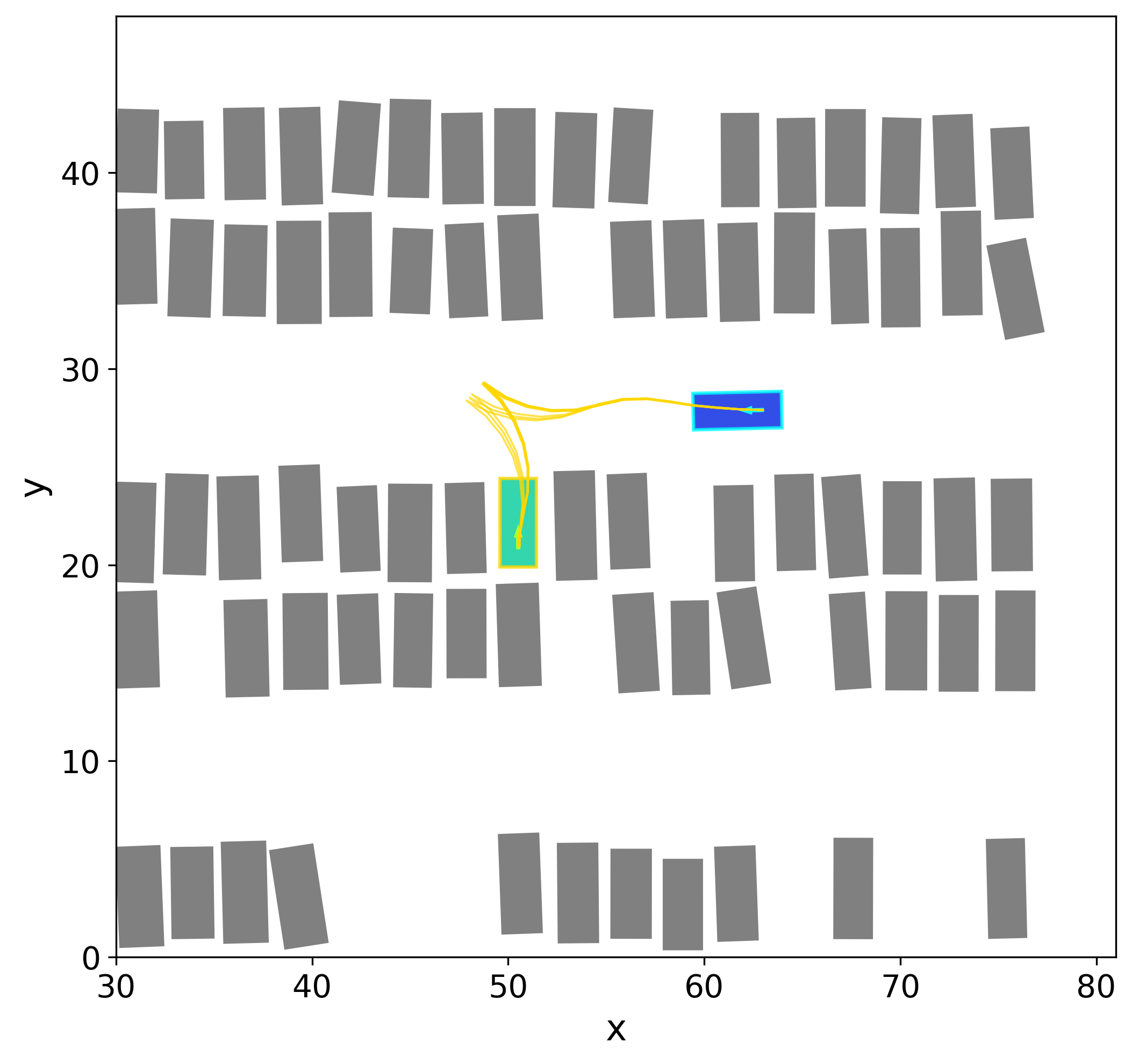}
  \end{minipage}}
\hfill
\subfloat[(c-2)]{%
  \begin{minipage}[t]{0.24\textwidth}\centering
    \includegraphics[height=\visfigheight]{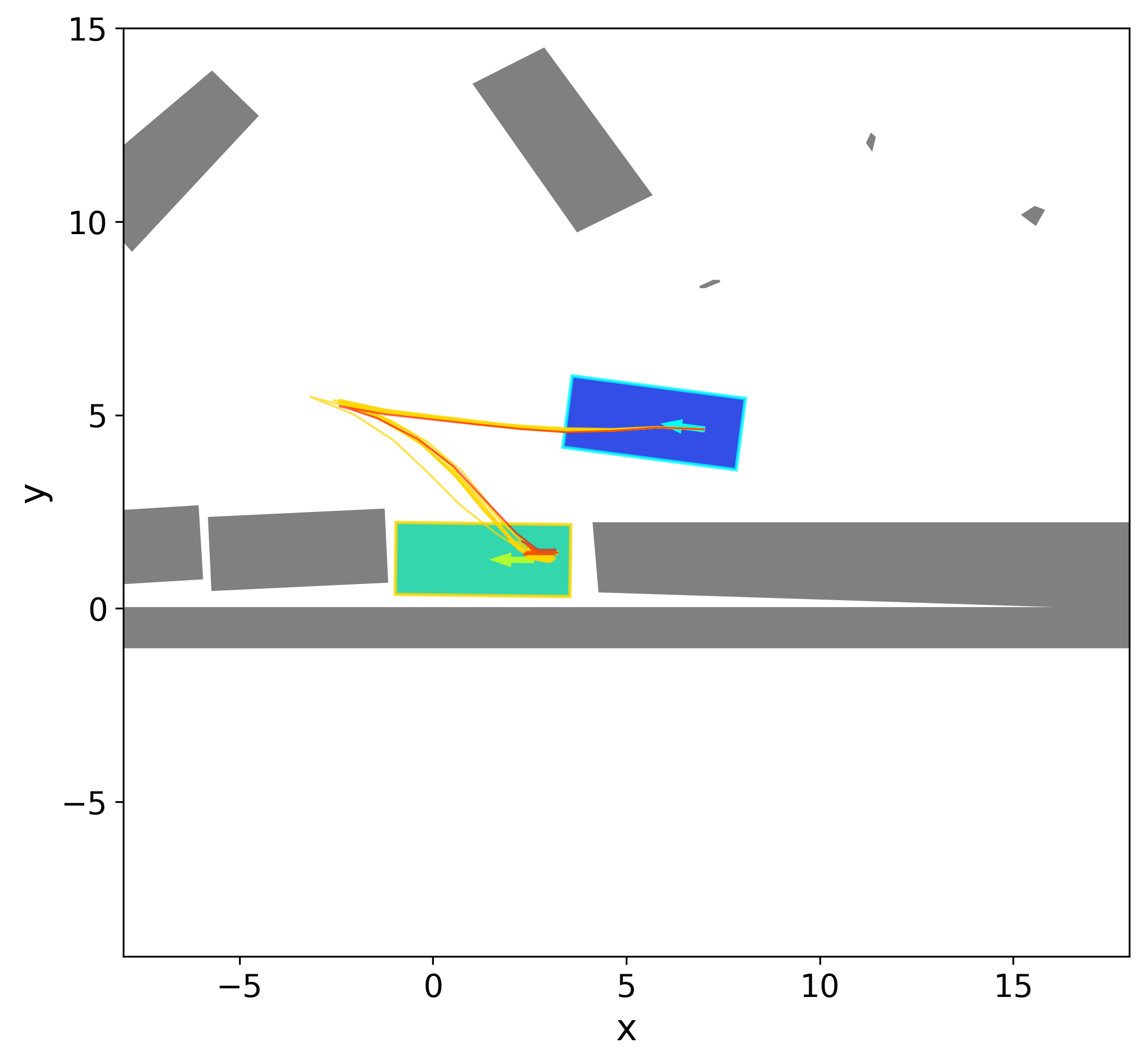}
  \end{minipage}}
\hfill
\subfloat[(d-2)]{%
  \begin{minipage}[t]{0.24\textwidth}\centering
    \includegraphics[height=\visfigheight]{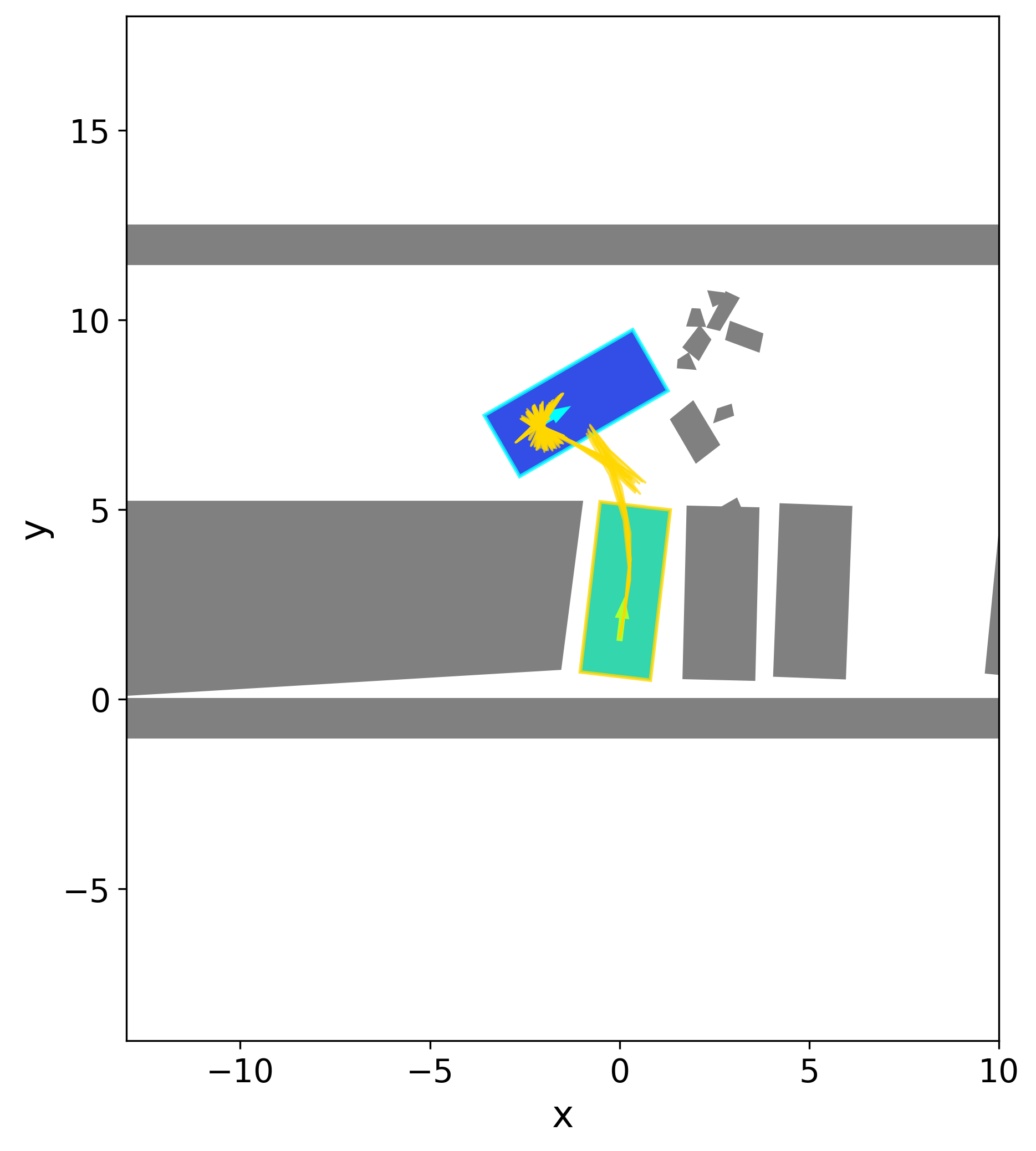}
  \end{minipage}}

\caption{Visualization for the comparison of planning results under 10 repeated trials per scenario. 
Yellow trajectories indicate successful runs that reach the target slot, while red trajectories indicate failures. 
\textbf{(a)} A normal vertical parking scenario with the RL baseline (a-1) and DRIP (a-2). 
\textbf{(b)} A complex vertical parking scenario reconstructed from the DLP dataset with the RL baseline (b-1) and DRIP (b-2). 
\textbf{(c)} A hard parallel parking scenario with the RL baseline (c-1) and DRIP (c-2). 
\textbf{(d)} A hard vertical parking scenario with the RL baseline (d-1) and DRIP (d-2).}
\vspace{5pt}

\label{fig:visualization}
\end{figure*}

\subsubsection{Trajectory quality metrics}

Beyond success rate, we evaluate DRIP against the SAC-based RL baseline from the perspective of execution cost and behavior quality. Table~\ref{tab:drip_vs_rl_with_delta} reports three metrics: (i) \emph{path length}, the accumulated Euclidean distance along the executed trajectory, reflecting travel efficiency and redundant maneuvering; (ii) \emph{interaction steps}, the number of planning iterations from the initial state to termination, indicating the efficiency of completing a task; and (iii) \emph{gear shifts}, the number of sign changes in consecutive velocity commands, measuring the frequency of switching between forward and reverse and thus the smoothness and practicality of the control sequence. We report both \textbf{Avg} (mean over repeated runs per scenario) and \textbf{Avg-Best} (best run per scenario, then averaged), characterizing robustness under stochasticity and the attainable performance under retries respectively.

Overall, DRIP consistently improves all three metrics. Compared with the RL baseline, DRIP reduces the overall Avg path length from 23.77\,m to 20.44\,m, interaction steps from 28.02 to 23.84, and gear shifts from 12.18 to 8.73, and the Avg-Best metrics show the same trend. The gains are modest in easy scenarios where the feasible action set is large, but become pronounced in medium and hard scenarios that require precise maneuvering under tight geometric constraints. Notably, improvements are generally larger on Avg than on Avg-Best, suggesting that DRIP primarily stabilizes and sharpens the policy’s output distribution, suppressing unnecessary adjustments rather than relying on occasional lucky trajectories.

\subsubsection{Qualitative visualization of planning results}
We further provide a qualitative comparison between the RL pre-trained baseline and the proposed diffusion refinement method to illustrate how DRIP improves robustness and maneuver quality. Fig.~\ref{fig:visualization} reports four representative scenarios, where each method is executed for 10 trials under the same setup.

In the normal vertical parking scenario (Fig.~\ref{fig:visualization} \textbf{(a)}), the RL baseline produces a visibly more dispersed trajectory set and can fail when the vehicle is blocked by nearby obstacles during reverse maneuvers. In contrast, DRIP achieves consistently successful runs with tighter trajectory overlap, suggesting improved stability. In the vertical parking scenario from DLP dataset (Fig.~\ref{fig:visualization} \textbf{(b)}), the RL baseline tends to switch to reversing early, which can yield an unfavorable entry angle and increase the likelihood of being blocked by nearby parked vehicles. DRIP more often preserves an appropriate forward probing distance before changing gear, resulting in a more reliable reverse entry.

The advantages of DRIP are more pronounced in more narrow scenarios. In the hard parallel parking scenario (Fig.~\ref{fig:visualization} \textbf{(c)}), where the longitudinal space between front and rear obstacles is extremely limited, the RL baseline succeeds 3 out of 10 trials, while DRIP substantially increases the success frequency and yields more consistent maneuver sequences, reaching a success rate of $9/10$. In the hard vertical parking scenario (Fig.~\ref{fig:visualization} \textbf{(d)}), RL often fails to align the vehicle with the narrow entrance within the step budget, whereas DRIP is able to execute feasible micro-adjustments and complete the park-in more reliably. Overall, these visualizations indicate that diffusion-based refinement concentrates the action distribution toward high-quality feasible solutions, with the largest gains in narrow and challenging parking geometries.

\newlength{\casefigheight}
\setlength{\casefigheight}{3.3cm}

\begin{figure*}[htb]
\centering
\captionsetup[subfigure]{labelformat=empty}
\subfloat[(a-1)]{%
  \begin{minipage}[t]{0.24\textwidth}\centering
    \includegraphics[height=3.1cm]{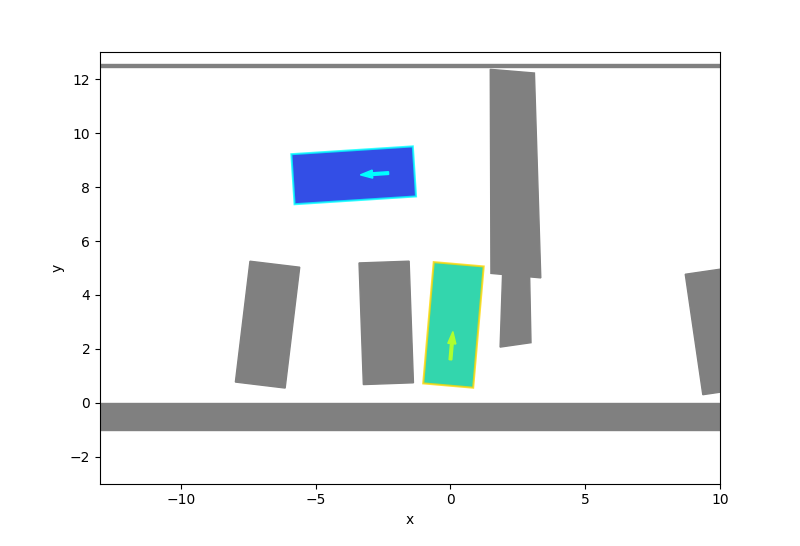}
  \end{minipage}}
\subfloat[(b-1)]{%
  \begin{minipage}[t]{0.24\textwidth}\centering
    \includegraphics[height=\casefigheight]{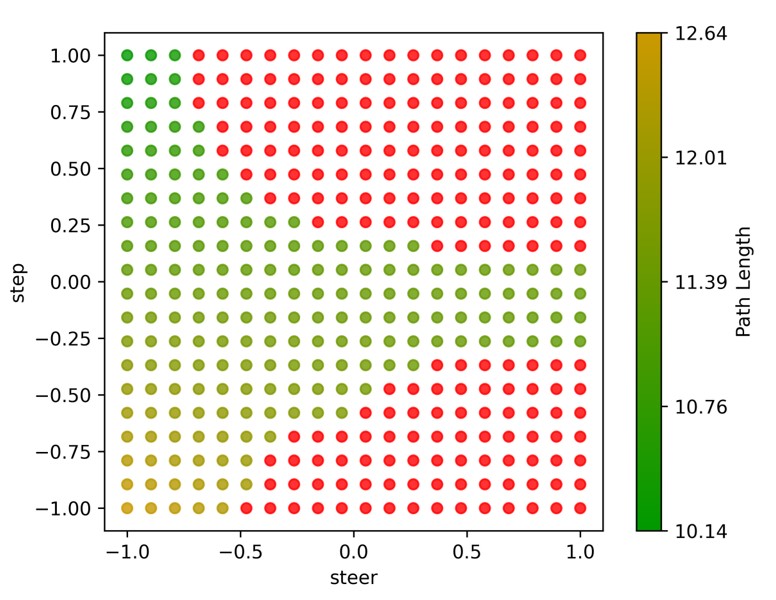}
  \end{minipage}}
  \hfill
\subfloat[(c-1)]{%
  \begin{minipage}[t]{0.24\textwidth}\centering
    \includegraphics[height=\casefigheight]{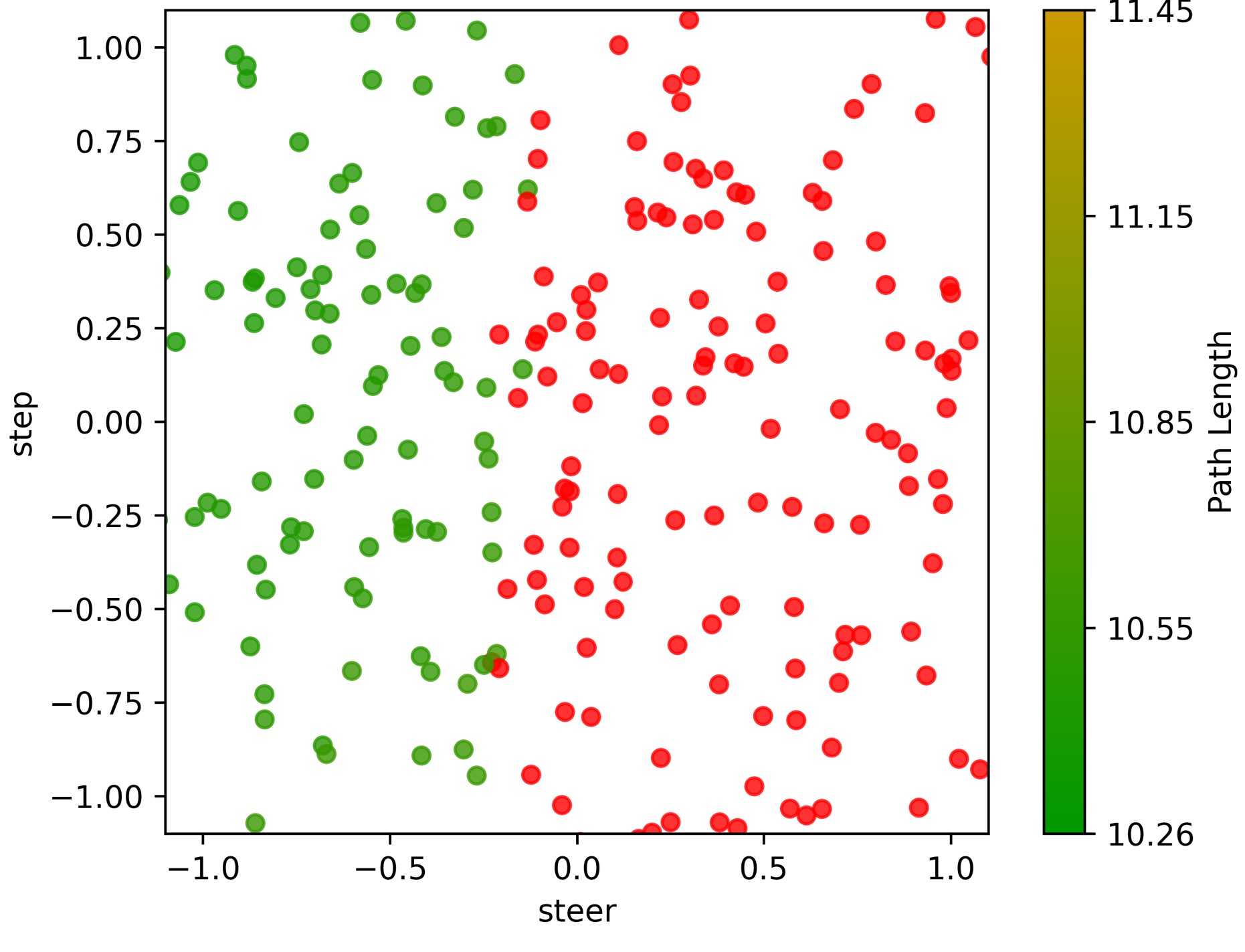}
  \end{minipage}}
  \hfill
\subfloat[(d-1)]{%
  \begin{minipage}[t]{0.24\textwidth}\centering
    \includegraphics[height=\casefigheight]{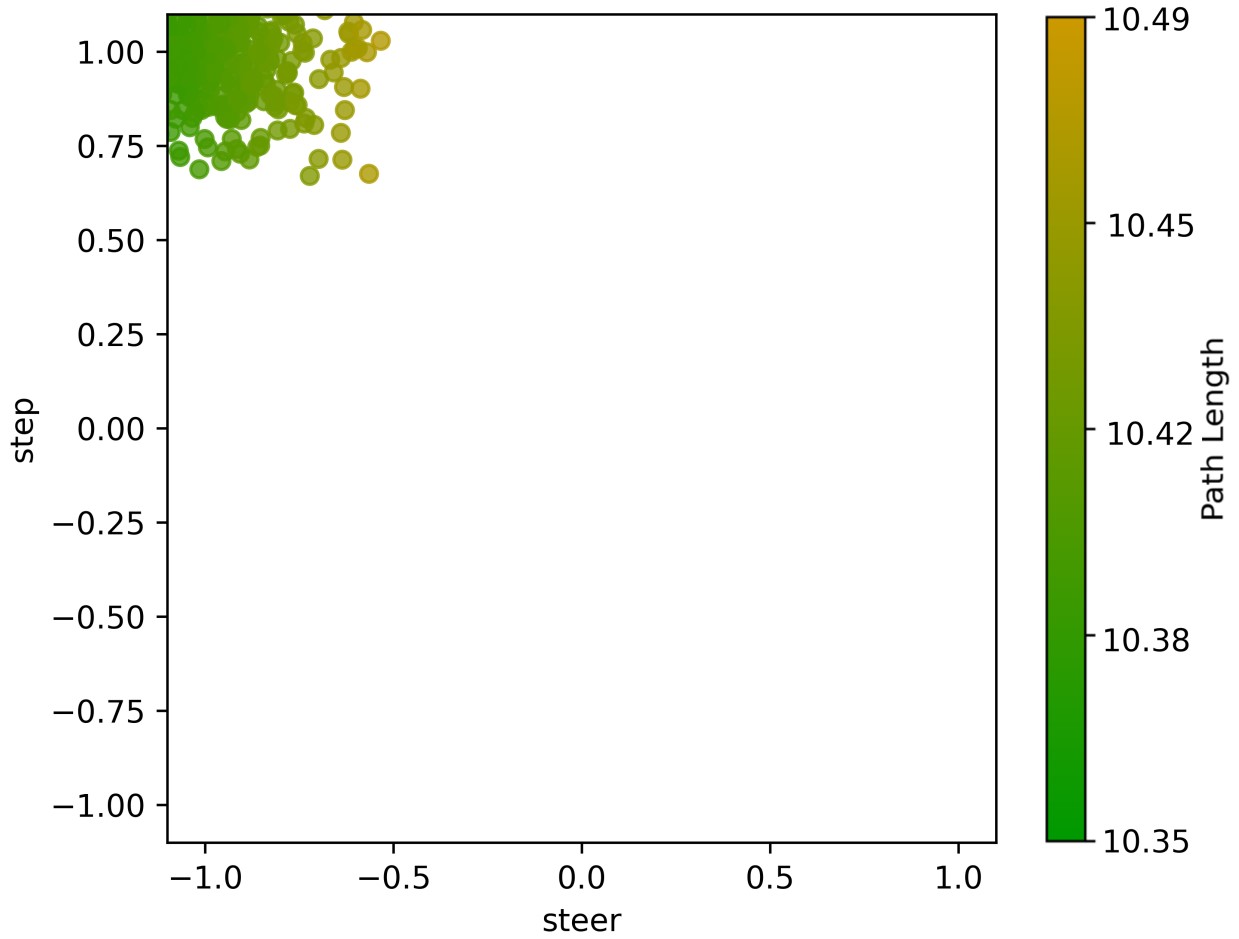}
  \end{minipage}}

\subfloat[(a-2)]{%
  \begin{minipage}[t]{0.24\textwidth}\centering
    \includegraphics[height=3.1cm]{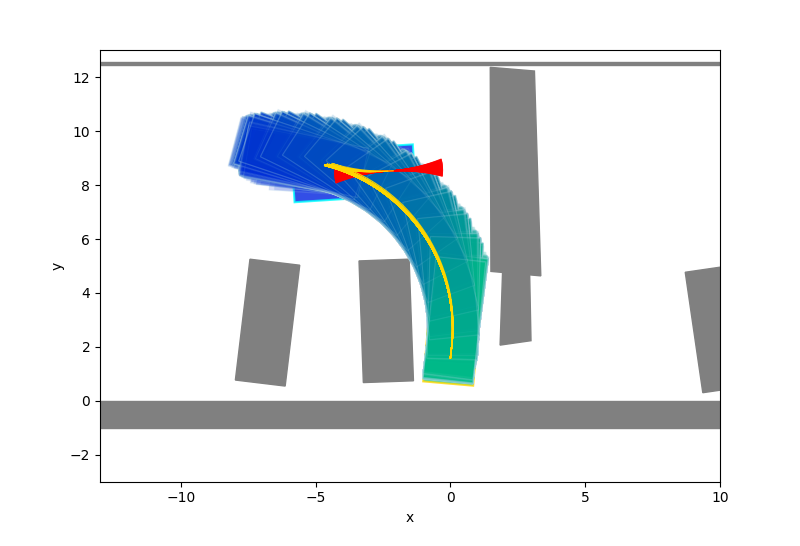}
  \end{minipage}}
\subfloat[(b-2)]{%
  \begin{minipage}[t]{0.24\textwidth}\centering
    \includegraphics[height=\casefigheight]{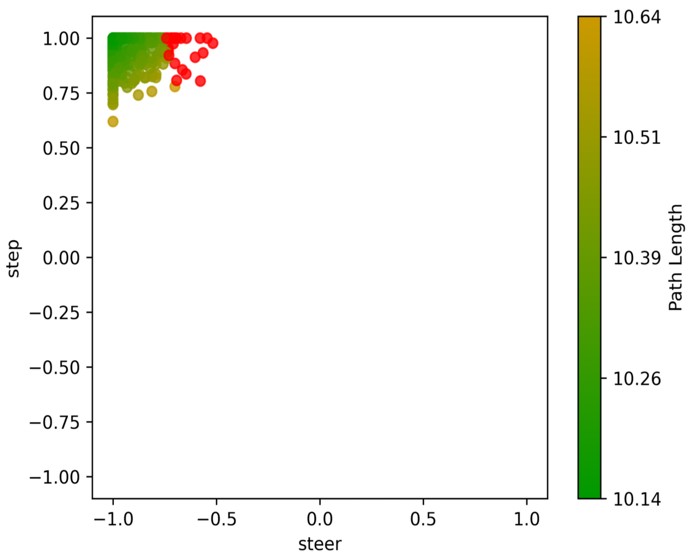}
  \end{minipage}}
  \hfill
\subfloat[(c-2)]{%
  \begin{minipage}[t]{0.24\textwidth}\centering
    \includegraphics[height=\casefigheight]{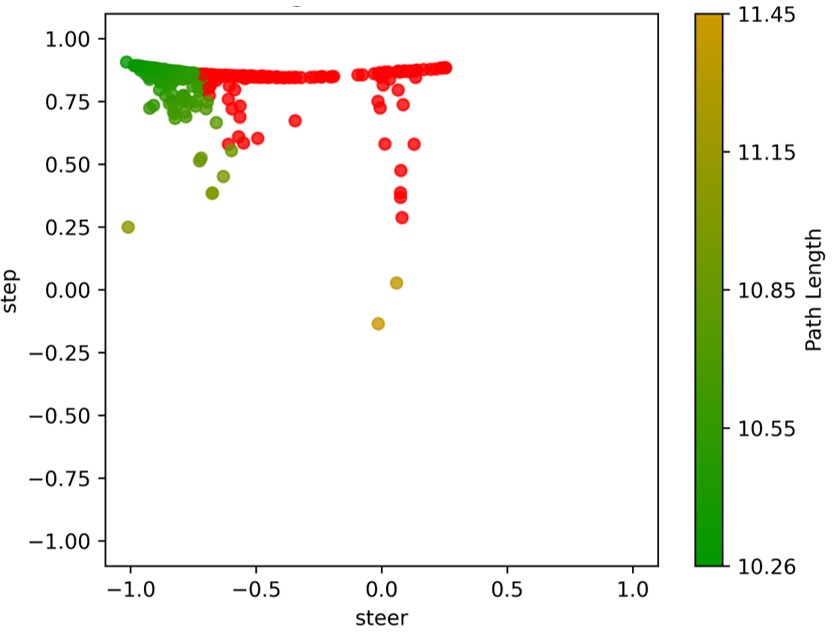}
  \end{minipage}}
  \hfill
\subfloat[(d-2)]{%
  \begin{minipage}[t]{0.24\textwidth}\centering
    \includegraphics[height=\casefigheight]{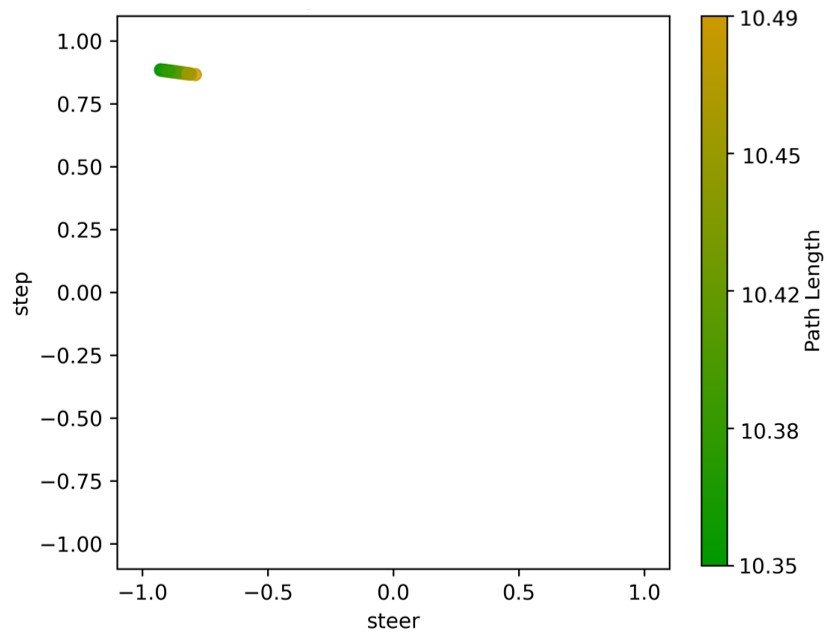}
  \end{minipage}}

\caption{Visualization for a vertical parking scenario.
\textbf{(a-1)} Scene overview.
\textbf{(a-2)} Feasible park-in trajectories (in yellow) and suboptimal actions (in red).
\textbf{(b-1)} Feasibility of 400 uniformly sampled actions in the action space; actions enabling a single-maneuver park-in with one gear change are highlighted in green/yellow. Other actions that require more than one gear change are in red.
\textbf{(b-2)} Samples from the RL-pretrained action distribution at the same position.
\textbf{(c-1)} Random action initialization from diffusion policy before denoising.
\textbf{(c-2)} Action distribution with diffusion policy after denoising.
\textbf{(d-1)} Proposed diffusion refinement initialized from the RL-pretrained prior distribution;
\textbf{(d-2)} Final action distribution after prior-aligned denoising.}
\label{fig: case study}
\end{figure*}

\subsubsection{Visualization for the denoising process}
We illustrate the effect of the proposed method with a case study in a vertical parking scene shown in Fig.~\ref{fig: case study}. As shown in Fig.~\ref{fig: case study} (a-1, a-2), a reasonable strategy is to move forward right and then the vehicle can complete parking with at most a single gear change. Actions that steer forward left, by contrast, often make a one-switch maneuver impossible, even at maximum steering the vehicle collides or becomes trapped by obstacles. To make this constraint explicit, we uniformly sample 400 actions in the local action space (step length, steering) and, for each action, check whether the post-action pose can be connected to the target slot by a Reeds-Shepp (RS) maneuver with one gear change \cite{reeds1990optimal}. Feasible actions are colored yellow/green in Fig.~\ref{fig: case study} (b-1), where greener indicates a shorter remaining path, while infeasible ones are red. The “good” actions concentrate in a narrow band in the upper-left region of the action space.

We then visualize three policies in the same position. The pretrained RL policy (Fig.~\ref{fig: case study} (b-2)) captures the overall structure of the "good" action distribution but still includes a nontrivial probability of suboptimal actions (about 4\% red in this case), reflecting the limits of parametric distribution modeling. A standard diffusion policy initialized from pure noise (Fig.~\ref{fig: case study} (c-1, c-2)) tends to denoise in the general direction towards the feasible band but struggles to concentrate probability within the green region. In contrast, our diffusion refinement starts denoising from the RL prior (Fig.~\ref{fig: case study} (d-1)). With prior-aligned training, it can concentrate the final distribution within the feasible band (Fig.~\ref{fig: case study} (d-2)), yielding a more precise action model for this challenging scenario.

\vspace{-5pt}
\subsection{Ablation Study.}

\renewcommand{\arraystretch}{1.15}
\begin{table}[t]
\centering
\caption{Ablation on prior-aligned training and truncation ratio $k^*/K$.}
\label{tab:ablation}
\small
\begin{tabular}{l|c|ccc|c}
\hline
\textbf{Method} & $\boldsymbol{k^*/K}$ & \textbf{Hard} & \textbf{Medium} & \textbf{Normal} & \textbf{Avg.} \\
\hline
\multicolumn{6}{c}{\textbf{Diffusion (IL)}} \\
\hline
DP & 1.0 & 48.3 & 85.2 & 92.7 & 76.8 \\
Prior init.\ only & 0.2 & 46.0 & 85.8 & 91.8 & 76.1 \\
DRIP  & 0.1 & 52.1 & 86.6 & 92.8 & 78.5 \\
DRIP (ours) & 0.2 & \textbf{59.1} & \textbf{88.9} & \textbf{96.2} & \textbf{82.4} \\
DRIP  & 0.4 & 46.8 & 83.4 & 88.8 & 74.5 \\
\hline
\multicolumn{6}{c}{\textbf{Diffusion (RL)}} \\
\hline
DQL & 1.0 & 45.0 & 86.4 & 91.0 & 75.9 \\
Prior init.\ only & 0.2 & 37.4 & 83.1 & 90.5 & 72.1 \\
DRIP  & 0.1 & 61.3 & 91.2 & 96.1 & 84.0 \\
DRIP (ours) & 0.2 & \textbf{65.9} & \textbf{92.0} & \textbf{96.8} & \textbf{85.9} \\
DRIP  & 0.4 & 54.1 & 91.7 & 94.6 & 81.8 \\
\hline
\end{tabular}
\end{table}

We evaluate the effect of the proposed prior-aligned training process. As shown in Tab.~\ref{tab:ablation}, directly using the RL-pretrained distribution as the initialization for truncated diffusion without alignment performs suboptimally relative to the corresponding DP/DQL baselines, and success rates exhibit slight drops. The reason is that, at the truncation step, the marginal of a standard diffusion process does not match the RL-pretrained distribution. Starting denoising from this mismatched prior leads to biased updates. In contrast, applying the proposed alignment during training yields consistent and significant improvements in success rate.

\begin{figure}[t]
    \centering
    \includegraphics[width=0.82\linewidth]{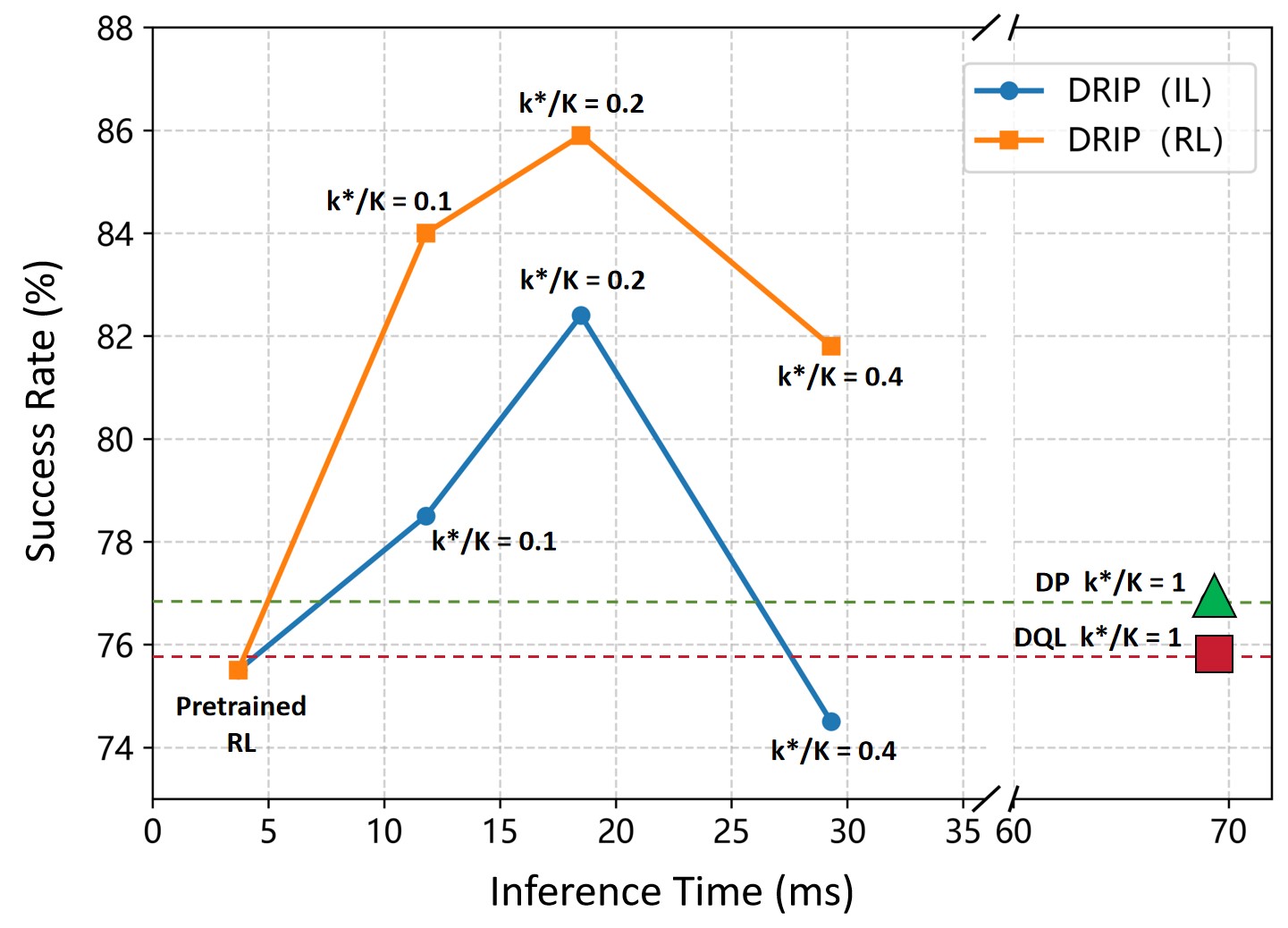}
    \caption{
        Comparison of average planning success rate (\%) and inference time (ms)
        between different methods.
    }
    \label{fig:success_vs_time}
\end{figure}

We select the truncation step \(k^{*}\) by Eq.~\eqref{eq: k selection}, i.e., the step whose standard-diffusion variance \(\sqrt{1-\bar{\alpha}_{k}}\) is closest to the RL prior’s standard deviation \(\sigma_{\mathrm{RL}}\). Therefore, initializing denoising at step $k^*$ provides the least biased inference, as the diffusion noise level at this step best matches the variance of the RL prior. In practice, with our pretrained policy and noise schedule, this gives \(k^{*}/K \approx 0.2\). With 25 inference steps, this corresponds to \(k^{*}=0.2\times 25=5\). Tab.~\ref{tab:ablation} and Fig.~\ref{fig:success_vs_time} also report results for alternative ratios \(k^{*}/K\). Both smaller and larger values degrade performance. Fewer denoising steps under-refine the action distribution, whereas more steps inject excessive noise that dilutes the prior. Besides, the selection of \(k^{*}/K = 0.2\) achieves the best performance, with a single-step inference time of 18.5 ms. This corresponds to a successful autoregressive generation of a full trajectory within one second, assuming the maximum number of steps is set to $50$. These observations further highlight the necessity of distributional alignment at the truncation point.

\subsection{Planning Case Study on Real-World Parking Scenarios}
\label{sec:real_world_parking}

To further examine the applicability of the proposed planner, we construct additional planning test cases from real-world underground parking scenarios. 
Specifically, the surrounding environment is captured by the LiDAR-based perception system of a real vehicle, and the resulting point cloud is projected and processed into a two-dimensional occupancy grid as the planner input, following the pipeline in our previous work~\cite{jiang2025hope}. 
We then evaluate the planning performance of DRIP and the RL baseline on these reconstructed grid maps.

\begin{figure}[h]
\centering
\captionsetup[subfigure]{labelformat=empty}
\newcommand{\realfigheight}{2.3cm}

\subfloat[(a-1) Scenario 1, RL]{%
  \begin{minipage}[t]{0.48\columnwidth}\centering
    \includegraphics[height=\realfigheight]{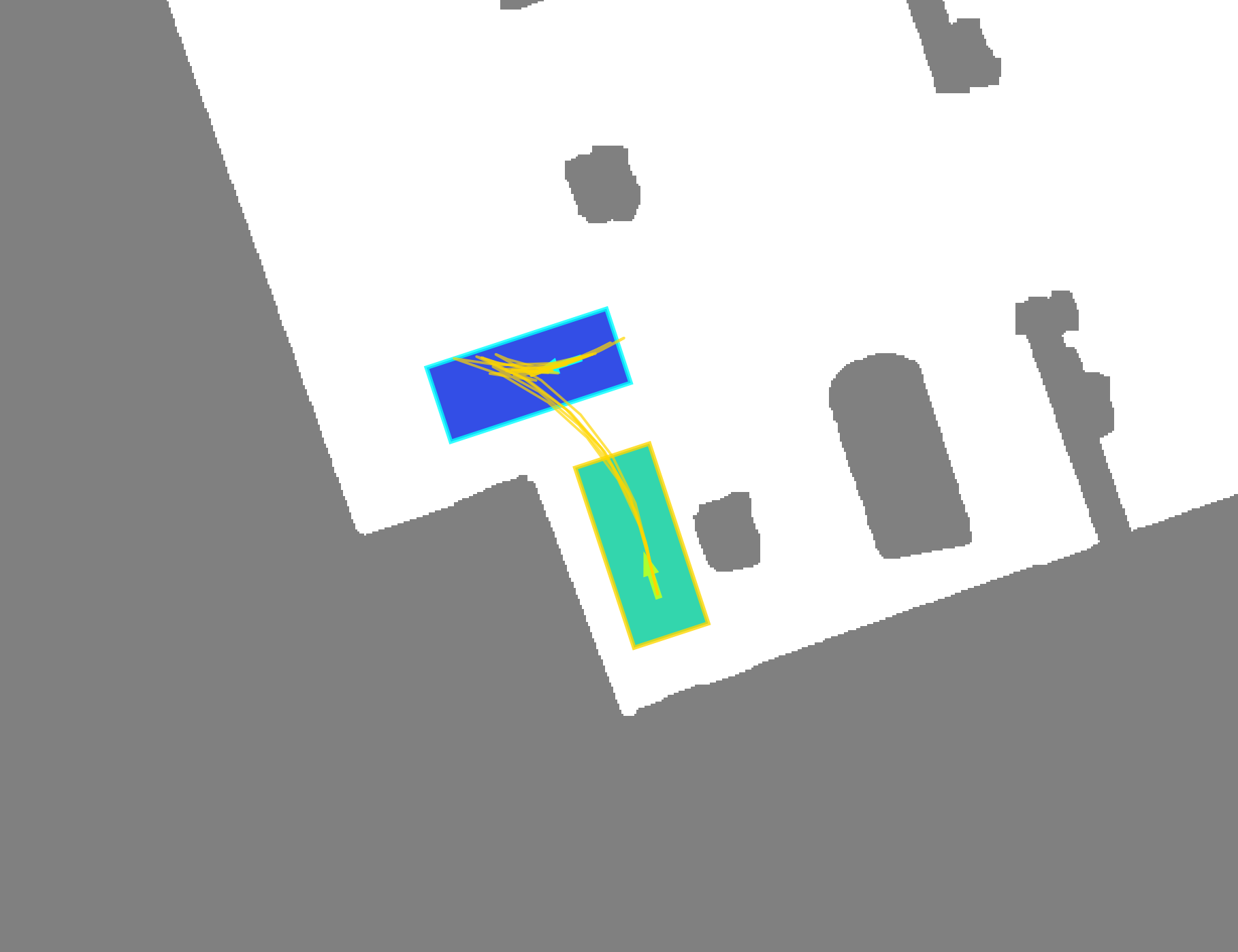}
  \end{minipage}}
\hfill
\subfloat[(b-1) Scenario 1, DRIP]{%
  \begin{minipage}[t]{0.48\columnwidth}\centering
    \includegraphics[height=\realfigheight]{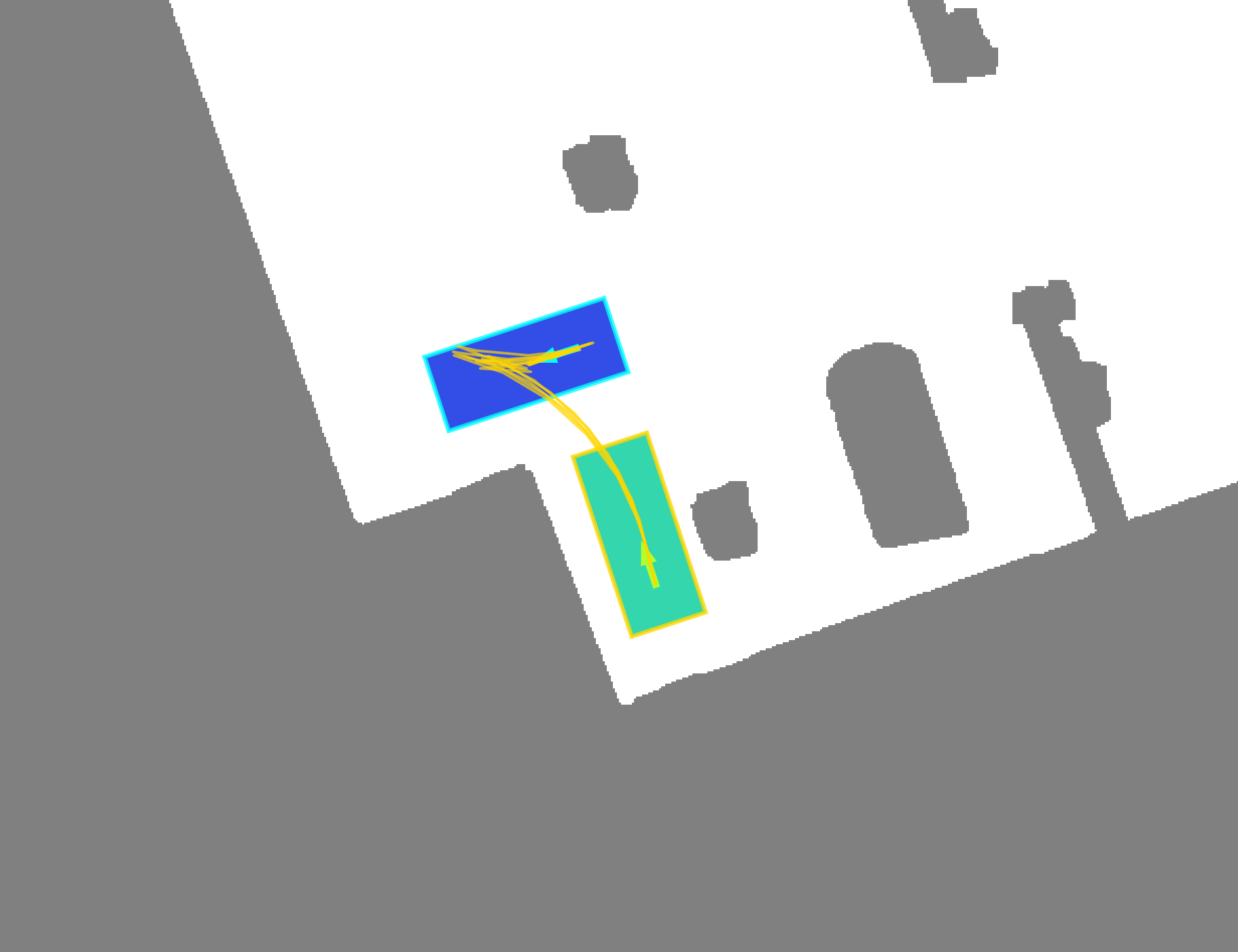}
  \end{minipage}}

\vspace{-0.7em}

\subfloat[(a-2) Scenario 2, RL]{%
  \begin{minipage}[t]{0.48\columnwidth}\centering
    \includegraphics[height=\realfigheight]{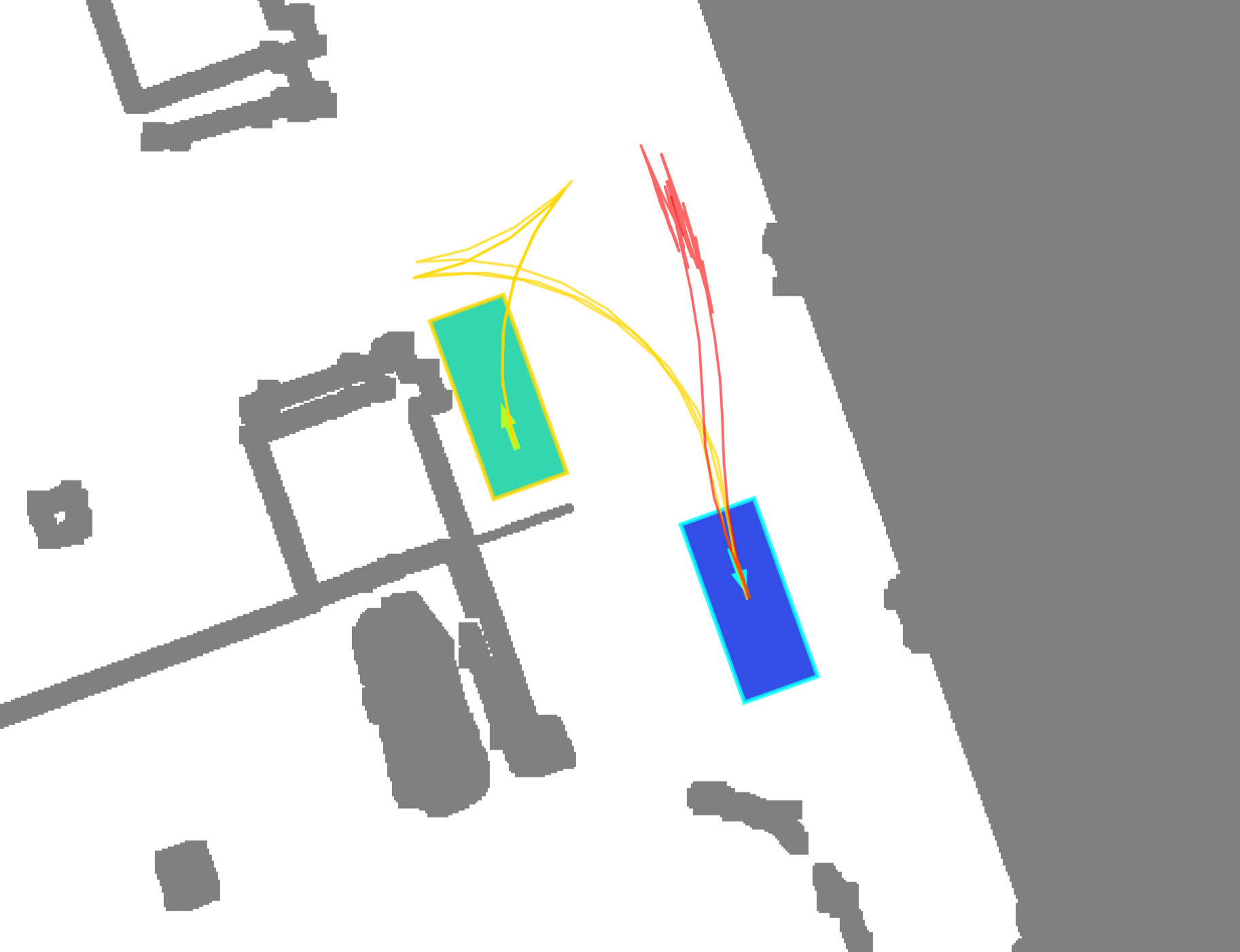}
  \end{minipage}}
\hfill
\subfloat[(b-2) Scenario 2, DRIP]{%
  \begin{minipage}[t]{0.48\columnwidth}\centering
    \includegraphics[height=\realfigheight]{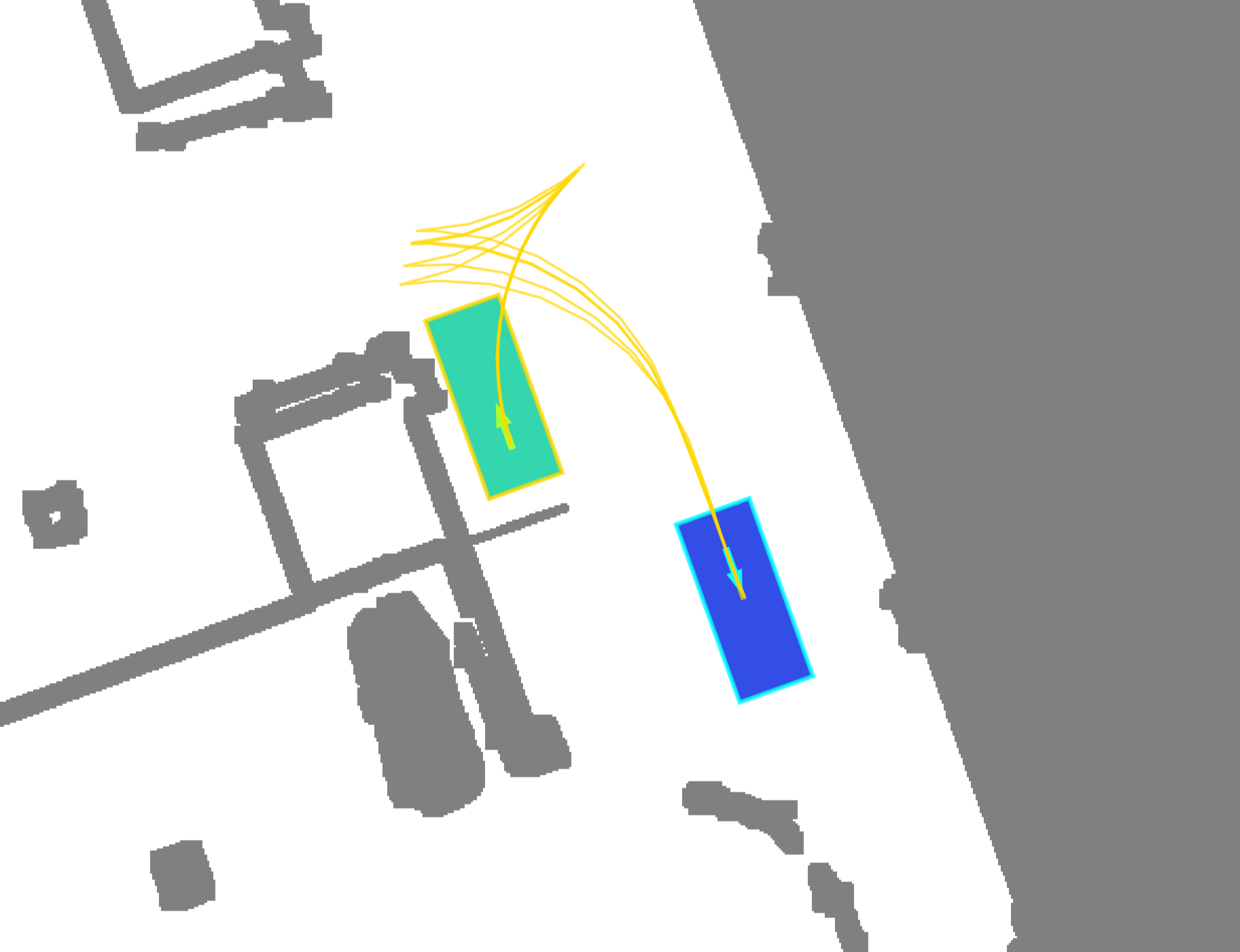}
  \end{minipage}}

\vspace{-0.7em}

\subfloat[(a-3) Scenario 3, RL]{%
  \begin{minipage}[t]{0.48\columnwidth}\centering
    \includegraphics[height=\realfigheight]{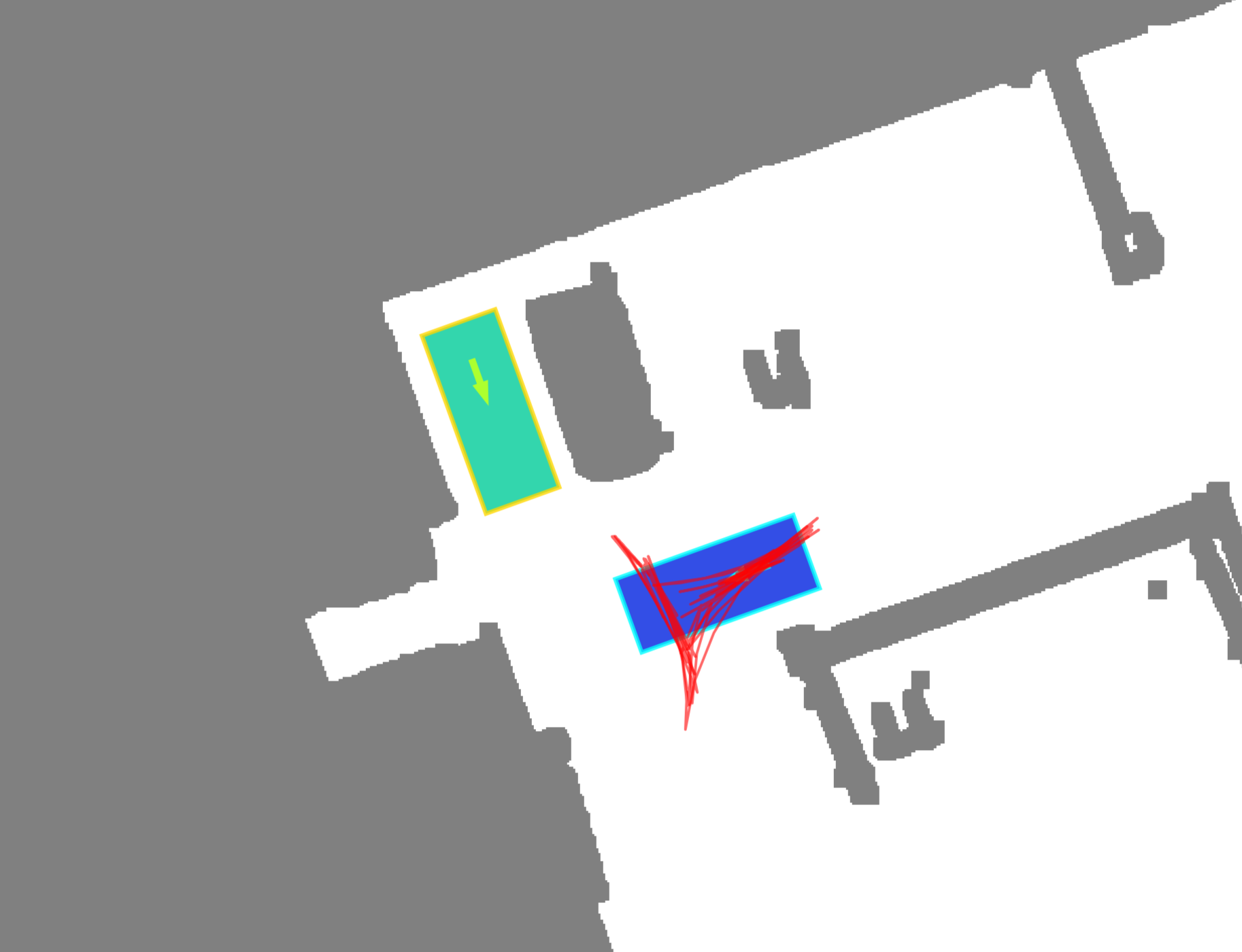}
  \end{minipage}}
\hfill
\subfloat[(b-3) Scenario 3, DRIP]{%
  \begin{minipage}[t]{0.48\columnwidth}\centering
    \includegraphics[height=\realfigheight]{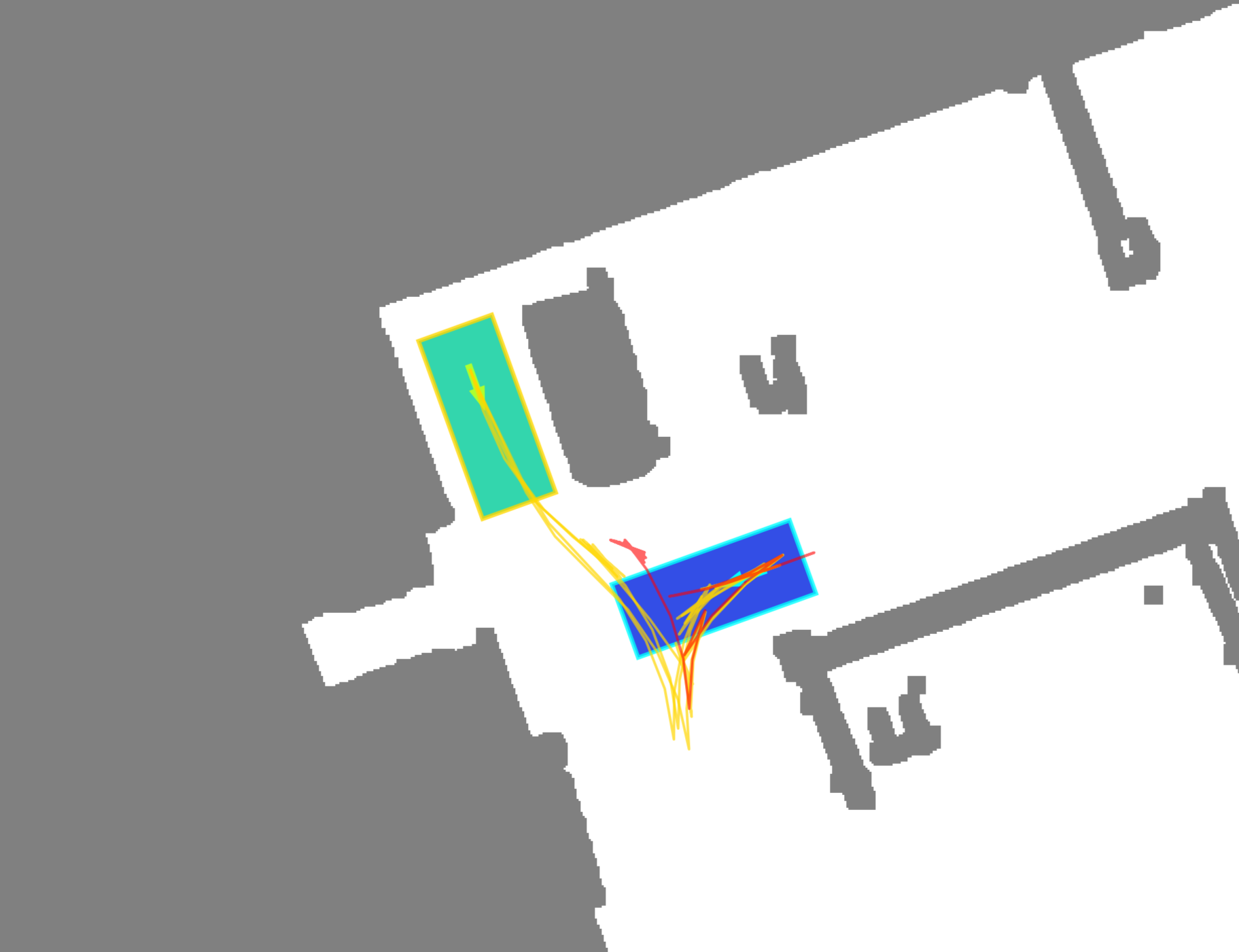}
  \end{minipage}}

\caption{Visualization in real-world reconstructed parking scenarios. Each row corresponds to one scenario, and the two columns compare the RL baseline and the proposed DRIP.}
\label{fig:real_world_case_study}
\end{figure}

To provide a focused scenario-level analysis, we consider three representative parking scenarios with different levels of maneuvering difficulty and repeat each test five times. 
The results are reported in Table~\ref{tab:real_world_scenarios}. 
Across these real-world reconstructed scenarios, DRIP achieves higher planning success and generates more efficient trajectories than the RL baseline. 
In the first perpendicular parking scenario , both methods successfully complete all trials, while DRIP requires fewer interaction steps, shorter path length, and fewer gear shifts. 
In the second scenario, where the parking slot direction is opposite to the initial vehicle heading and the vehicle needs to reorient before parking, the RL baseline sometimes produces imprecise initial maneuvers, causing the vehicle to move close to the wall and making subsequent parking more difficult. 
By contrast, DRIP achieves stable success in all trials. 
The advantage of DRIP becomes more pronounced in the third scenario, where the parking slot is close to the wall and a protruding obstacle exists in front of the slot. 
This setting requires precise adjustment of the vehicle pose during reversing, under which DRIP maintains a high success rate, whereas the RL baseline fails in all trials.

\begin{table}[t]
\centering
\caption{Planning performance on real-world LiDAR-based gridmap.}
\label{tab:real_world_scenarios}
\resizebox{\columnwidth}{!}{
\begin{tabular}{c|c|c|c|c|c}
\toprule
\textbf{Scenario} & \textbf{Method} & \textbf{Success} & \textbf{Path Len. (m) $\downarrow$} & \textbf{Steps $\downarrow$} & \textbf{Gear Shifts $\downarrow$} \\
\midrule
\multirow{2}{*}{1} 
& RL   & \textbf{5/5} & 16.03 & 21.8 & 8.6 \\
& DRIP & \textbf{5/5} & \textbf{13.56} & \textbf{16.4} & \textbf{5.4} \\
\midrule
\multirow{2}{*}{2} 
& RL   & 3/5 & 28.82 & 33.8 & 13.4 \\
& DRIP & \textbf{5/5} & \textbf{24.18} & \textbf{22.2} & \textbf{2.6} \\
\midrule
\multirow{2}{*}{3} 
& RL   & 0/5 & 46.26 & 51.0 & 30.4 \\
& DRIP & \textbf{4/5} & \textbf{29.61} & \textbf{38.6} & \textbf{19.2} \\
\bottomrule
\end{tabular}
}
\end{table}

\section{Conclusion and Future Work}

This work presented DRIP, a diffusion-refined RL planner for space-constrained automated parking. 
Motivated by the need for reliable maneuver planning in space-constrained parking environments, where feasible trajectories occupy a limited region of the action space, we propose to refine RL-pretrained action distributions through diffusion-based denoising. 
Specifically, an RL-pretrained policy first provides a structured maneuver prior, and a diffusion model then refines this prior into more precise and feasible parking actions. 
A key technical challenge is that standard diffusion training produces intermediate marginals whose mean and variance generally do not match the pretrained prior at the truncation point, resulting in a training-inference mismatch when inference is initialized from the pretrained distribution. 
To address this issue, we introduced a prior-aligned training procedure that modifies the forward noising process with a time-varying mean, ensuring that the denoising trajectory is distributionally aligned with the RL prior at a chosen step. 
This design allows inference to start from the pretrained action distribution and perform a short, well-informed denoising process for high-precision parking planning.

Across the parking benchmark, including \emph{normal}, \emph{medium}, and \emph{hard} scenes, and additional real-world reconstructed parking scenarios, DRIP consistently improves the RL-pretrained action distribution and achieves higher planning success rates. 
Compared with diffusion policies trained without RL priors, DRIP also obtains significant performance gains, especially in the hard and most space-constrained scenarios. 
Visualization, ablation, and scenario-level case studies further indicate that the improvement stems not only from using a prior distribution but critically from the proposed prior alignment. Mere parameter initialization is insufficient to realize these benefits.

Future work will extend DRIP toward practical parking-related applications, including  autonomous valet parking and intelligent parking-lot management. 
In these scenarios, reliable planning in space-constrained environments is essential for coordinating vehicle maneuvers, reducing unnecessary movements, and improving parking-space utilization. 
By generating more precise and feasible trajectories under tight spatial constraints, DRIP has the potential to support safer and more efficient low-speed mobility in dense parking environments, contributing to reduced collision risk and improved parking efficiency in future transportation systems.

\bibliography{ref}
\bibliographystyle{IEEEtran}

\end{document}